\documentclass[11pt, a4paper, onecolumn, copyright, goog]{handshaketemplate}

\usepackage[
  backend=biber,
  style=authoryear,
  natbib=true,
  maxcitenames=2,
  maxbibnames=99,
  uniquelist=false,
  uniquename=false,
  dashed=false,
  mincrossrefs=99,
]{biblatex}
\DeclareDelimFormat{nameyeardelim}{\addcomma\space}
\DeclareDelimFormat{multinamedelim}{\addsemicolon\space}
\DeclareDelimFormat{finalnamedelim}{\addspace\&\space}
\DeclareDelimFormat{multicitedelim}{\addsemicolon\space}

\AtEveryBibitem{\clearfield{urldate}\clearfield{urlyear}\clearfield{urlmonth}\clearfield{urlday}}

\newcommand{\ishandshake}{}

\usepackage{bm}

\ifdefined\ishandshake
  \newcommand{\hbm}[1]{\textbf{#1}}
\else
  \newcommand{\hbm}[1]{$#1$}
\fi
\usepackage{multirow}
\usepackage{lineno}
\usepackage{wrapfig}
\usepackage{silence}
\usepackage{tikz}
\usetikzlibrary{arrows.meta}
\usepackage{pgfplots}
\usepackage{pgfplotstable}
\usepgfplotslibrary{groupplots,fillbetween}
\pgfplotsset{compat=1.18}
\usepackage{standalone}

\usepackage{csquotes}

\definecolor{darkblue}{rgb}{0, 0, 0.5}
\hypersetup{colorlinks=true, citecolor=darkblue, linkcolor=darkblue, urlcolor=darkblue}

\newcommand{\cornerlogo}[1][1]{%
  \begin{tikzpicture}[remember picture,overlay]
    \node[anchor=north west, xshift=2.0cm, yshift=-1.8cm] at (current page.north west)
      {\includegraphics[scale=#1]{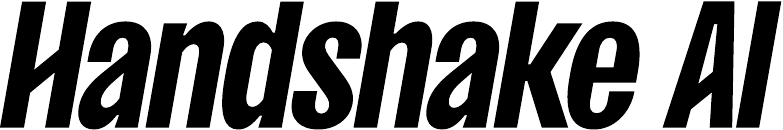}};
    \draw[line width=0.3pt]
      ($(current page.north west)+(2.0cm,-2.7cm)$) -- ($(current page.north east)+(-2.0cm,-2.7cm)$);
  \end{tikzpicture}%
}

\title{An Imperfect Verifier is Good Enough: Learning with Noisy Rewards}

\renewcommand{\today}{}

\makeatletter
\renewcommand\AB@authnote[1]{\textsuperscript{#1}}
\renewcommand\AB@affilnote[1]{\textsuperscript{#1}}
\makeatother

\author[1,2]{Andreas Plesner}
\author[1]{Francisco Guzm\'{a}n}
\author[1]{Anish Athalye}
\affil[1]{Handshake AI}
\affil[2]{ETH Zurich}

\begin{abstract}
Reinforcement Learning with Verifiable Rewards (RLVR) is widely used for post-training Large Language Models, but practical verifiers can make errors. We study how the rate and structure of reward noise affect RLVR in code generation, with a preliminary scientific-reasoning check.
In multi-seed Qwen3~8B experiments on MBPP, mean validation reward over two fixed late evaluations is within 1 percentage point of the clean baseline at the tested resampled group-rollout noise rates through \hbm{20\%}, and within about 2 points at 30\%. Confidence intervals allow larger losses; these point estimates do not establish a general tolerance threshold. We also examine full-program pass@k, four controlled noise structures, two model-based verifiers, and policy models from three families spanning 4B--9B parameters.
We derive conditional advantage distributions for symmetric and asymmetric group noise, including retained format penalties, and show why clipping limits simple gradient-scaling arguments. The analysis identifies information preserved by whole-group corruption and limits on interpreting our asymmetric sweep as a precision--recall comparison.
Overall, the results indicate that imperfect verification can support effective RLVR in the tested settings, while aggregate error rates alone do not characterize the learning signal.
 \end{abstract}

\begin{document}

\cornerlogo[0.27]

\maketitle

\section{Introduction}

Reinforcement learning (RL) has grown in popularity as a post-training method to improve large language models (LLMs) in various domains, particularly after the release of DeepSeek-R1~\citep{deepseek-aiDeepSeekR1IncentivizingReasoning2025}, which demonstrated that Reinforcement Learning with Verifiable Rewards (RLVR) and Group Relative Policy Optimization (GRPO)~\citep{shaoDeepSeekMathPushingLimits2024} can produce a frontier-level model at relatively low cost. Early efforts focused on verifiable domains such as math and coding, where deterministic checks provide the reward signal.
Since then, RLVR has been extended to semi-verifiable domains such as finance and law, where rubrics and LLM-as-a-Judge provide the reward signal~\citep{viswanathanChecklistsAreBetter2025,zhouBreakingExplorationBottleneck2025}.

RLVR is motivated by the ideal of a \textit{perfect} verifier: a deterministic oracle that consistently rewards good outputs and penalizes bad ones. In practice, though, such a verifier does not exist.
While some datasets, such as GPQA (answers are multiple-choice) and AIME (answers are integers between 000 and 999), admit highly accurate verifiers, verification can be challenging and error-prone even in ostensibly verifiable domains such as mathematics. For example, a string-equality-based verifier might incorrectly reject $\frac{1}{e^2} - \frac{1}{6}$ as an answer with respect to a ground truth $\frac{6 - e^2}{6e^2}$, despite mathematical equivalence~\citep{xuTinyVReducingFalse2025,huang:rl-verifier-robustness}. This problem is exacerbated in semi-verifiable domains where non-deterministic model-based judges are used~\citep{tan:judgebench,pan:rubriceval}.

Verifier accuracy has improved through domain-specific techniques, such as Math-Verify~\citep{huggingface:math-verify}, and fine-tuned judges~\citep{zhu:judgelm,heAdvancedIFRubricBasedBenchmarking2025}. A practical question remains: \textit{under which error structures and training conditions can an imperfect verifier support effective RLVR?} This question also matters for LLM-as-a-Judge and Agent-as-a-Judge~\citep{vidgenAPEXAgents2026,zhuge:agent-as-a-judge}, where errors may depend systematically on the generated response.

To study these questions in a controlled setting, we focus on RLVR for coding, and we measure the impact of verifier noise on model training. We focus on coding for two reasons: because the reward structure mirrors rubric-based rewards (a set of unit tests is analogous to a set of text-based rubric criteria), and because the verifiers can be highly accurate, providing a noise-free baseline as a point of comparison for controlled and realistic noise.

Our multi-seed Qwen3~8B experiments on MBPP show small observed losses under resampled group-rollout noise: the mean over two fixed late evaluations is within $1$ percentage point of the clean baseline through $20\%$ noise, and within about $2$ points at $30\%$. The confidence intervals allow larger losses, so these point estimates do not establish a one-point tolerance guarantee. We also compare noise structures and two model-based verifiers, and include a preliminary scientific-reasoning experiment. Our analysis explains why whole-group corruption preserves useful advantage information in expectation and why our asymmetric design cannot isolate a general precision--recall trade-off.

\section{Related work}
\label{sec:relwk}

\paragraph{Reinforcement learning with verifiable rewards.}
\citet{silverWelcomeEraExperience2025} articulated a vision for scaling RL to language agents, crystallizing an emerging trend of increasing model abilities through RL~\citep{openaiLearningReasonLLMs2024,zengSimpleRLZooInvestigatingTaming2025}. A key example of this is RLVR, which was popularized by DeepSeek-R1~\citep{deepseek-aiDeepSeekR1IncentivizingReasoning2025}. They showed that GRPO with outcome-based rewards on math and code tasks can produce strong reasoning LLMs while having less risk of reward hacking than learned reward models~\citep{gaoScalingLawsReward2023}. 
Verification can also be applied at the process level rather than just the outcome to provide stronger learning signals~\citep{lightmanLetsVerifyStep2023,hubotterReinforcementLearningSelfDistillation2026}. 
Our work takes RLVR as a given (see \Cref{app: background} or \citet{deepseek-aiDeepSeekR1IncentivizingReasoning2025} for background) and investigates what happens when the outcome-based reward signal is imperfect.

\paragraph{LLM-as-a-Judge / model-based verifiers.}
Using models as evaluators has become standard practice for evaluation and training~\citep{baiConstitutionalAIHarmlessness2022,guSurveyLLMasaJudge2026}. \citet{zhengJudgingLLMasajudgeMTbench2023a} introduced the LLM-as-a-Judge paradigm for evaluation, and \citet{lambertRewardBenchEvaluatingReward2025} created a benchmark to rank judges' alignment with human preferences.
\citet{liuExaminingReasoningLLMsasJudges2026} show that reasoning models are better verifiers than non-reasoning models.
For training, rubric-based approaches use model-based verifiers to provide reward signals in domains without deterministic verifiers~\citep{gunjalRubricsRewardsReinforcement2025, heAdvancedIFRubricBasedBenchmarking2025}. 

Model-based verifiers have well-documented failure modes, including positional bias~\citep{thakurJudgingJudgesEvaluating2025} and sensitivity to prompt phrasing~\citep{chenHumansLLMsJudge2024,shankarWhoValidatesValidators2024}, meaning that model-based verifier noise is neither uniform nor independent.

\paragraph{Reinforcement learning with noisy rewards.}
The problem of learning under corrupted/noisy rewards is not new. \citet{everittReinforcementLearningCorrupted2017} formalize the corrupted reward channel as a POMDP~\citep{kaelblingPlanningActingPartially1998} and identify conditions under which optimal behavior is still recoverable. \citet{wangReinforcementLearningPerturbed2020} propose estimators robust to reward perturbation in deep RL. In the preference-based setting, \citet{gongAdaptiveConfidenceawarePreferencebased2025} study noisy feedback and \citet{liEvaluatingFeatureDependent2026} analyze feature-dependent noise. Separately, \citet{fortunatoNoisyNetworksExploration2018} add parametric noise directly to network weights to drive exploration.

\paragraph{Noisy rewards in RLVR.}
\citet{radRateFateRLVepsilonR2026} analyze reward noise in GRPO and validate their analysis on Qwen2.5~3B for code generation. Their prediction that positively informative noise can primarily affect learning speed is consistent with our observations; our study adds comparisons across noise structures and model families at finite training budgets.
\citet{caiReinforcementLearningVerifiable2025} study asymmetric synthetic and model-based verifier noise in mathematical reasoning and evaluate reward corrections. Differences in task and evaluation protocol limit direct numerical comparisons with our results (\Cref{sec:cai protocol comparison}).
\citet{shaoSpuriousRewardsRethinking2025} study spurious rewards in RLVR for math, finding that Qwen models can learn from incorrect reward signals. \citet{zhuNoisyDataDestructive2026} follow up with a rebuttal saying that noise does hurt model performance. \citet{chenExplorationVsExploitation2026} revisit spurious rewards through the lens of exploration vs. exploitation, and \citet{xuExplorationExploitationTwoStage2025} propose a two-stage entropy-based approach for noise-tolerant RLVR training.

These studies cover different corruption mechanisms and noise levels. We focus on observed performance under specified noise structures. \citet{mansouriNoisecorrectedGRPONoisy2025} model Bernoulli reward corruption and derive a gradient correction. We evaluate uncorrected training; without a correction baseline, our experiments do not determine when such methods would improve performance.

\section{Methodology}
\label{sec:noise}

We begin with a noise-free baseline of RLVR in the coding domain, where the verifier executes the supplied unit tests. We check that the ground truth solutions pass all tests. Throughout, ``error-free'' refers to these test outcomes; a finite test suite does not establish program correctness on every input.

Next, we train models using noisy verifiers, evaluating the trained model against the held-out validation split \emph{using the error-free verifier}. For the verifier used during training, we vary noise rates as well as noise \emph{type}: we study a more controlled setting of specific noise patterns as well as more realistic noise distributions by using a model-based verifier.

\subsection{Controlled noise for RLVR}\label{sec: controlled noise}

In the code-generation setting we study, each prompt has $T$ unit tests, and each training step generates a group of $G$ rollouts per prompt. Test outcomes are represented by a $G \times T$ binary matrix $\mathbf{M}$, where $M_{i,j}=1$ if rollout $i$ passes test $j$; the test-based reward is the row mean. Execution failure receives a clean reward of $0$. Missing a fenced Python code block instead receives a format penalty of $-0.25$, which is retained unchanged under corruption and is excluded from the test-outcome model below.

We inject noise by flipping entries of $\mathbf{M}$ with probability $p$, independently of whether the original value is pass or fail. This bitflipping means the false positive rate and false negative rate are both equal to $p$. We identify two orthogonal axes that determine the \emph{structure} of the noise, yielding four modes (illustrated in \Cref{fig:noise-overview}):

\begin{enumerate}
    \item[\textbf{(a)}] \textbf{Sample $\times$ unit test.} Each cell $M_{i,j}$ is flipped independently with probability $p$. This noise is the most fine-grained; every test outcome for every rollout has an independent chance of corruption.
    \item[\textbf{(b)}] \textbf{Sample $\times$ rollout.} Each row (rollout) is selected with probability $p$, and all test outcomes in that row are flipped. This introduces perfectly correlated errors across tests within a rollout.
    \item[\textbf{(c)}] \textbf{Group $\times$ unit test.} Each column (test index) is selected with probability $p$, and that test's outcome is flipped for \emph{all} rollouts in the group. This introduces perfectly correlated errors across rollouts for a test within the sampled group.
    \item[\textbf{(d)}] \textbf{Group $\times$ rollout.} The entire matrix is flipped with probability $p$. When triggered, every test outcome for every rollout is inverted; otherwise, the matrix is left intact. This noise is the coarsest; it either corrupts everything or nothing.
\end{enumerate}

\begin{figure}[t]
    \centering
    \begin{tikzpicture}[
  font=\sffamily\small,
  brace/.style={
    decorate,
    decoration={calligraphic brace, amplitude=4pt, raise=2pt},
    very thick,
  },
]

  \def\cs{0.40}
  \def\nr{8}
  \def\nc{3}
  \pgfmathsetmacro{\mw}{\nc*\cs}
  \pgfmathsetmacro{\mh}{\nr*\cs}
  \def\noisyX{3.2}
  \def\hgap{3.2}
  \def\groupsep{4.6} %

  \definecolor{flip}{RGB}{255,235,238}
  \definecolor{flipborder}{RGB}{183,28,28}
  \definecolor{passgreen}{RGB}{46,125,50}
  \definecolor{failgray}{RGB}{140,140,140}
  \definecolor{hdrcolor}{RGB}{55,71,79}

  \newcommand{\passsymbol}{{\bfseries\ding{51}}}
  \newcommand{\failsymbol}{{\bfseries\ding{55}}}

  \newcommand{\drawgrid}{%
    \draw[gray!50, thin] (0,0) grid[step=\cs] (\mw,\mh);
    \draw[thick] (0,0) rectangle (\mw,\mh);
  }

  \newcommand{\drawmatrix}[1]{%
    \foreach \r/\c/\v/\f in {#1} {%
      \pgfmathtruncatemacro{\isflip}{\f}%
      \ifnum\isflip=1\relax
        \fill[flip] ({(\c-1)*\cs},{(\nr-\r)*\cs}) rectangle +(\cs,\cs);
      \fi
    }%
    \drawgrid
    \foreach \r/\c/\v/\f in {#1} {%
      \pgfmathtruncatemacro{\ispass}{\v}%
      \pgfmathtruncatemacro{\isflip}{\f}%
      \ifnum\isflip=1\relax
        \colorlet{cellcolor}{flipborder}%
      \else
        \ifnum\ispass=1\relax
          \colorlet{cellcolor}{passgreen}%
        \else
          \colorlet{cellcolor}{failgray}%
        \fi
      \fi
      \ifnum\ispass=1\relax
        \node[font=\footnotesize, text=cellcolor]
          at ({(\c-1)*\cs+\cs/2},{(\nr-\r)*\cs+\cs/2}) {\passsymbol};
      \else
        \node[font=\footnotesize, text=cellcolor]
          at ({(\c-1)*\cs+\cs/2},{(\nr-\r)*\cs+\cs/2}) {\failsymbol};
      \fi
    }%
  }

  \pgfmathsetmacro{\subtitleY}{\mh+0.15}
  \pgfmathsetmacro{\ruleY}{\mh+0.65}
  \pgfmathsetmacro{\titleY}{\mh+0.75}

  \begin{scope}[shift={(0,0)}]
    \drawmatrix{%
      1/1/1/0, 1/2/0/0, 1/3/1/0,
      2/1/1/0, 2/2/1/0, 2/3/0/0,
      3/1/0/0, 3/2/1/0, 3/3/1/0,
      4/1/1/0, 4/2/0/0, 4/3/1/0,
      5/1/0/0, 5/2/1/0, 5/3/0/0,
      6/1/1/0, 6/2/1/0, 6/3/1/0,
      7/1/0/0, 7/2/0/0, 7/3/1/0,
      8/1/1/0, 8/2/1/0, 8/3/0/0%
    }
    \foreach \r in {1,...,8} {
      \node[font=\sffamily\tiny, anchor=east, text=gray]
        at (-0.10, {(\nr-\r)*\cs+\cs/2}) {$r_{\r}$};
    }
    \foreach \c in {1,2,3} {
      \node[font=\sffamily\tiny, anchor=south, text=gray]
        at ({(\c-1)*\cs+\cs/2}, {\mh+0.08}) {$t_{\c}$};
    }
  \node[font=\sffamily\bfseries, text=hdrcolor, anchor=south] at ({\mw/2}, \ruleY) {Original};
  \end{scope}

  \draw[-{Stealth[length=3mm]}, thick, hdrcolor]
    ({\mw+0.30}, {\mh/2}) -- ({\noisyX-0.30}, {\mh/2})
    node[midway, above=2pt, font=\sffamily\scriptsize, text=hdrcolor] {$+$\,noise};

  \pgfmathsetmacro{\sL}{\noisyX}
  \pgfmathsetmacro{\sR}{\noisyX+\hgap+\mw}
  \node[font=\sffamily\bfseries, text=hdrcolor, anchor=south]
    at ({(\sL+\sR)/2}, \titleY) {Sample level};

  \pgfmathsetmacro{\gL}{\noisyX+\hgap+\groupsep}
  \pgfmathsetmacro{\gR}{\noisyX+2*\hgap+\groupsep+\mw}
  \node[font=\sffamily\bfseries, text=hdrcolor, anchor=south]
    at ({(\gL+\gR)/2}, \titleY) {Group level};

  \pgfmathsetmacro{\sepX}{(\sR+\gL)/2}
  \draw[hdrcolor!40, dashed]
    (\sepX, -0.15) -- (\sepX, {\titleY-0.1});

  \begin{scope}[shift={(\noisyX,0)}]
    \drawmatrix{%
      1/1/1/0, 1/2/1/1, 1/3/1/0,
      2/1/0/1, 2/2/1/0, 2/3/1/1,
      3/1/0/0, 3/2/1/0, 3/3/0/1,
      4/1/1/0, 4/2/0/0, 4/3/1/0,
      5/1/0/0, 5/2/0/1, 5/3/0/0,
      6/1/0/1, 6/2/1/0, 6/3/0/1,
      7/1/1/1, 7/2/0/0, 7/3/1/0,
      8/1/1/0, 8/2/0/1, 8/3/0/0%
    }
    \node[font=\sffamily\footnotesize, text=hdrcolor, anchor=south]
      at ({\mw/2}, \subtitleY) {(a) Unit test independent};
    \node[font=\sffamily\scriptsize, text=gray, anchor=north, align=center]
      at ({\mw/2}, {-0.2}) {Each cell\\with probability $p$};
  \end{scope}

  \pgfmathsetmacro{\bX}{\noisyX+\hgap}
  \begin{scope}[shift={(\bX,0)}]
    \drawmatrix{%
      1/1/1/0, 1/2/0/0, 1/3/1/0,
      2/1/0/1, 2/2/0/1, 2/3/1/1,
      3/1/0/0, 3/2/1/0, 3/3/1/0,
      4/1/1/0, 4/2/0/0, 4/3/1/0,
      5/1/1/1, 5/2/0/1, 5/3/1/1,
      6/1/1/0, 6/2/1/0, 6/3/1/0,
      7/1/1/1, 7/2/1/1, 7/3/0/1,
      8/1/1/0, 8/2/1/0, 8/3/0/0%
    }

    \node[font=\sffamily\footnotesize, text=hdrcolor, anchor=south]
      at ({\mw/2}, \subtitleY) {(b) Entire rollout};
    \node[font=\sffamily\scriptsize, text=gray, anchor=north, align=center]
      at ({\mw/2}, {-0.20}) {Each row\\with probability $p$};
  \end{scope}

  \pgfmathsetmacro{\cX}{\noisyX+\hgap+\groupsep}
  \begin{scope}[shift={(\cX,0)}]
    \drawmatrix{%
      1/1/0/1, 1/2/0/0, 1/3/0/1,
      2/1/0/1, 2/2/1/0, 2/3/1/1,
      3/1/1/1, 3/2/1/0, 3/3/0/1,
      4/1/0/1, 4/2/0/0, 4/3/0/1,
      5/1/1/1, 5/2/1/0, 5/3/1/1,
      6/1/0/1, 6/2/1/0, 6/3/0/1,
      7/1/1/1, 7/2/0/0, 7/3/0/1,
      8/1/0/1, 8/2/1/0, 8/3/1/1%
    }

    \node[font=\sffamily\footnotesize, text=hdrcolor, anchor=south]
      at ({\mw/2}, \subtitleY) {(c) Unit test independent};
    \node[font=\sffamily\scriptsize, text=gray, anchor=north, align=center]
      at ({\mw/2}, {-0.20}) {Each column\\with probability $p$};
  \end{scope}

  \pgfmathsetmacro{\dX}{\noisyX+2*\hgap+\groupsep}
  \begin{scope}[shift={(\dX,0)}]
    \drawmatrix{%
      1/1/0/1, 1/2/1/1, 1/3/0/1,
      2/1/0/1, 2/2/0/1, 2/3/1/1,
      3/1/1/1, 3/2/0/1, 3/3/0/1,
      4/1/0/1, 4/2/1/1, 4/3/0/1,
      5/1/1/1, 5/2/0/1, 5/3/1/1,
      6/1/0/1, 6/2/0/1, 6/3/0/1,
      7/1/1/1, 7/2/1/1, 7/3/0/1,
      8/1/0/1, 8/2/0/1, 8/3/1/1%
    }

    \node[font=\sffamily\footnotesize, text=hdrcolor, anchor=south]
      at ({\mw/2}, \subtitleY) {(d) Entire rollout};
    \node[font=\sffamily\scriptsize, text=gray, anchor=north, align=center]
      at ({\mw/2}, {-0.20}) {Entire matrix\\with probability $p$};
  \end{scope}

  \pgfmathsetmacro{\legendY}{-1.50}
  \pgfmathsetmacro{\legendX}{(\noisyX+\dX+\mw)/2}
  \node[font=\footnotesize, text=passgreen, anchor=base] at ({\legendX-2.0}, {\legendY}) {\passsymbol};
  \node[font=\sffamily\scriptsize, text=hdrcolor, anchor=base west]
    at ({\legendX-1.87}, {\legendY}) {pass};
  \node[font=\footnotesize, text=failgray, anchor=base] at ({\legendX-0.80}, {\legendY}) {\failsymbol};
  \node[font=\sffamily\scriptsize, text=hdrcolor, anchor=base west]
    at ({\legendX-0.73}, {\legendY}) {fail};
  \fill[flip] ({\legendX+0.25}, {\legendY-0.05}) rectangle +(0.25,0.25);
  \draw[thick] ({\legendX+0.25}, {\legendY-0.05}) rectangle +(0.25,0.25);
  \node[font=\sffamily\scriptsize, text=hdrcolor, anchor=base west]
    at ({\legendX+0.62}, {\legendY}) {flipped};

\end{tikzpicture}
    \caption{The four controlled noise injection modes. Rows are rollouts ($r_1$--$r_8$), columns are unit tests ($t_1$--$t_3$). Red cells indicate flipped outcomes. \textbf{(a)}~Each cell flipped independently. \textbf{(b)}~Entire rows flipped. \textbf{(c)}~Entire columns flipped. \textbf{(d)}~Entire matrix flipped.}
    \label{fig:noise-overview}
\end{figure}

Noise is resampled independently on each encounter with a prompt across training epochs. The main sweep uses equal false-negative and false-positive rates; an exploratory asymmetric sweep appears in \Cref{app:additional-results}. These controlled experiments do not model persistent verifier blind spots. \Cref{app:group-noise-analysis} derives the group-noise advantage distributions, including retained format penalties and the limits of gradient identities under clipping.

\subsection{Model-based verifier}

Controlled corruption specifies error rates and correlations but does not capture all verifier failure modes. We therefore also replace the unit-test executor with a model-based verifier that predicts whether each test would pass given the generated code. It receives the code and unit tests as assert statements and returns binary judgments, aggregated into a per-rollout reward (\Cref{sec:prompts}). Comparing two verifier models changes both the rate and structure of their errors; it is not an intervention on error rate alone.

This setup resembles rubric-based post-training~\citep{gunjalRubricsRewardsReinforcement2025, heAdvancedIFRubricBasedBenchmarking2025}, where a judge grades individual criteria and aggregates them into a score. Here, executing the tests provides reference labels for computing verifier accuracy, precision, recall, and F1 throughout training. These metrics describe each judge on the evolving outputs of its own trained policy; they do not compare judges on a shared, fixed response distribution. Predicting test execution also need not reproduce the error structure of rubric judgments in other domains.

\section{Experimental setup}
\label{sec: experimental setup}

\paragraph{Dataset.} Our experiments focus on the Mostly Basic Python Problems~(MBPP) dataset~\citep{austinProgramSynthesisLarge2021}. Each problem provides a natural-language description and three unit tests. We train on the standard MBPP training split (374 problems) and evaluate on the validation split (90 problems) using a maximum response length of 8192 tokens. We report mean unit-test reward, including format penalties, and pass@k, where a successful program must pass all three tests (\Cref{app:additional-results}).

\paragraph{Models and compute.} We use an internal training framework built on SLIME~\citep{slime_github}. For our main experiments, we focus on Qwen3~8B~\citep{yangQwen3TechnicalReport2025} and GLM4~9B (\texttt{GLM-Z1-9B-0414})~\citep{glm2024chatglmfamilylargelanguage}, with ablations using Llama~3.1~8B and Qwen3~4B. Training uses Group Relative Policy Optimization~(GRPO)~\citep{shaoDeepSeekMathPushingLimits2024}; full hyperparameters are in \Cref{sec:training-details}.

Each training run requires approximately 64 GPU-hours on H100 GPUs. Multi-seed experiments are therefore limited to the most important configurations to manage compute costs. Unless otherwise noted, results are from a single seed. The expanded Qwen3~8B group-rollout sweep uses 3--6 seeds per setting, including 4 baseline seeds; pass@k uses 2--3 seeds. Other configurations retain the seed counts shown in \Cref{tab: overview}.

\section{Results}
We first report the group-rollout noise sweep, then compare the four controlled noise types at $p{=}0.10$ and the two model-based verifiers. The comparisons concern the recorded training budgets and validation problems.

\subsection{Observed tolerance to group-rollout noise}

The recorded group-rollout sweeps cover noise rates up to $p{=}0.50$. For the expanded Qwen3~8B cohort, the mean of evaluation steps 239 and 259 (\Cref{tab:fixed-checkpoints}) is within $1$ percentage point of the clean baseline ($0.890$) at the tested rates $p{=}0.05$--$0.20$, with observed losses of $1.4$ points at $p{=}0.25$ and $2.0$ points at $p{=}0.30$. Uncertainty is wider than the one-point margin: the 95\% interval for the difference at $p{=}0.20$ is $[-2.9,+1.8]$ points, and at $p{=}0.05$ it is $[-4.3,+2.7]$. Peak scores show a similar pattern (\Cref{fig:noise-sweep-best}), but include unequal checkpoint-selection opportunities. Performance drops substantially at $p{=}0.40$ and above. At $p{=}0.50$, the test-outcome signal is uninformative in expectation before normalization, but the unchanged format penalties still provide a signal.

\Cref{tab:rebuttal-peak,tab:rebuttal-passk} report the expanded results. For peak unit-test rewards, two-sided independent equal-variance $t$-tests against the clean baseline give unadjusted $p$-values of $0.309$ at $15\%$ noise, $0.028$ at $25\%$, and $0.023$ at $30\%$; a nonsignificant result does not establish equivalence. Peak pass@1 is $0.816$ at $15\%$ noise versus $0.814$ for its two-seed baseline. At $40\%$, pass@1 falls to $0.756$ even though pass@16 matches the baseline, illustrating that tolerance depends on the evaluation metric. We complement peak scores with final-checkpoint results and training curves (\Cref{fig:noise-sweep-final,app: time to best}).

\begin{figure}[t]
    \centering
    \begin{tikzpicture}
\pgfplotsset{table/search path={.,shared/diagram/noise_sweep,shared/results/rebuttal,../../results/rebuttal,shared/results/training,../../results/training}}
\pgfplotstableread[col sep=comma]{arxiv-shared--results--training--peak_sweep_glm.csv}\sweepdata
\pgfplotstableread[col sep=comma]{arxiv-shared--results--rebuttal--peak_scores_summary.csv}\expandeddata

\pgfplotstablegetelem{0}{mean}\of{\sweepdata}
\edef\glmBaselineBest{\pgfplotsretval}
\pgfplotstablegetelem{0}{mean_peak}\of{\expandeddata}
\edef\qwenBaselineBest{\pgfplotsretval}

\begin{groupplot}[
    group style={group size=2 by 1, horizontal sep=1.8cm},
    width=0.55\columnwidth,
    height=0.45\columnwidth,
    xmin=0.0, xmax=0.5,
    ymin=0.45, ymax=0.95,
    xlabel={Noise level ($p$)},
    ylabel={Validation Reward},
    xtick={0.0,0.1,0.2,0.3,0.4,0.5},
    xticklabels={0.00,0.10,0.20,0.30,0.40,0.50},
    ytick={0.50,0.60,0.70,0.80,0.90},
    yticklabels={0.50,0.60,0.70,0.80,0.90},
    ymajorgrids=true,
    unbounded coords=jump,
    filter discard warning=false,
    grid style={dashed,gray!35},
]

\nextgroupplot[
    title={GLM4 9B},
    legend style={at={(0.03,0.03)},anchor=south west,draw=black,fill=white},
]
\addplot[name path=glmBestUpper, draw=none, forget plot] table[x=noise_p, y expr={\thisrow{n}>1 ? \thisrow{mean}+\thisrow{sd_plot} : nan}] {\sweepdata};
\addplot[name path=glmBestLower, draw=none, forget plot] table[x=noise_p, y expr={\thisrow{n}>1 ? \thisrow{mean}-\thisrow{sd_plot} : nan}] {\sweepdata};
\addplot[blue!35, fill opacity=0.750, forget plot] fill between[of=glmBestUpper and glmBestLower];
\addplot[color=blue!75!black, mark=*, mark size=2.2pt, line width=1.8pt, forget plot]
    table[x=noise_p, y=mean] {\sweepdata};
\addlegendimage{
    legend image code/.code={
        \draw[blue!75!black, line width=1.8pt] (0cm,0cm) -- (0.5cm,0cm);
        \fill[blue!75!black] (0.25cm,0cm) circle (1.6pt);
    }
}
\addlegendentry{Group Rollout Noise}
\addplot[black, dashed, line width=1.0pt, forget plot] coordinates {(-0.1,\glmBaselineBest) (0.6,\glmBaselineBest)};
\addlegendimage{
    legend image code/.code={
        \draw[black, dashed, line width=1.0pt] (0cm,0cm) -- (0.5cm,0cm);
    }
}
\addlegendentry{No Noise (\pgfmathprintnumber[fixed,precision=3]{\glmBaselineBest})}

\nextgroupplot[
    title={Qwen3 8B},
    legend style={at={(0.03,0.03)},anchor=south west,draw=black,fill=white},
]
\addplot[name path=qwenBestUpper, draw=none, forget plot] table[x=noise_probability, y expr=\thisrow{mean_peak}+\thisrow{sd_peak}] {\expandeddata};
\addplot[name path=qwenBestLower, draw=none, forget plot] table[x=noise_probability, y expr=\thisrow{mean_peak}-\thisrow{sd_peak}] {\expandeddata};
\addplot[orange!45, fill opacity=0.750, forget plot] fill between[of=qwenBestUpper and qwenBestLower];
\addplot[color=orange!90!black, mark=*, mark size=2.2pt, line width=1.8pt, forget plot]
    table[x=noise_probability, y=mean_peak] {\expandeddata};
\addlegendimage{
    legend image code/.code={
        \draw[orange!90!black, line width=1.8pt] (0cm,0cm) -- (0.5cm,0cm);
        \fill[orange!90!black] (0.25cm,0cm) circle (1.6pt);
    }
}
\addlegendentry{Group Rollout Noise}
\addplot[black, dashed, line width=1.0pt, forget plot] coordinates {(-0.1,\qwenBaselineBest) (0.6,\qwenBaselineBest)};
\addlegendimage{
    legend image code/.code={
        \draw[black, dashed, line width=1.0pt] (0cm,0cm) -- (0.5cm,0cm);
    }
}
\addlegendentry{No Noise (\pgfmathprintnumber[fixed,precision=3]{\qwenBaselineBest})}

\end{groupplot}
\end{tikzpicture}
    \caption{Best validation reward across noise levels for group rollout noise. Shaded regions indicate $\pm 1$ standard deviation across seeds. Qwen3~8B uses the expanded 3--6-seed results, including later checkpoints where available; GLM4~9B uses recovered peaks through rollout step 259. These peak results include different run lengths; \Cref{fig:noise-sweep-final} compares a common rollout step.}
    \label{fig:noise-sweep-best}
\end{figure}

\begin{figure}[t]
    \centering
    \begin{tikzpicture}
\pgfplotsset{table/search path={.,shared/results/training,../../results/training}}
\pgfplotstableread[col sep=comma]{arxiv-shared--results--training--final_sweep_glm.csv}\glmfinaldata
\pgfplotstableread[col sep=comma]{arxiv-shared--results--training--final_sweep_qwen.csv}\qwenfinaldata

\pgfplotstablegetelem{0}{mean}\of{\glmfinaldata}
\edef\glmBaselineFinal{\pgfplotsretval}
\pgfplotstablegetelem{0}{mean}\of{\qwenfinaldata}
\edef\qwenBaselineFinal{\pgfplotsretval}

\begin{groupplot}[
    group style={group size=2 by 1, horizontal sep=1.8cm},
    width=0.55\columnwidth,
    height=0.45\columnwidth,
    xmin=0.0, xmax=0.5,
    ymin=-0.10, ymax=0.95,
    xlabel={Noise level ($p$)},
    ylabel={Validation Reward},
    xtick={0.0,0.1,0.2,0.3,0.4,0.5},
    xticklabels={0.00,0.10,0.20,0.30,0.40,0.50},
    ytick={0.0,0.2,0.4,0.6,0.8},
    yticklabels={0.0,0.2,0.4,0.6,0.8},
    ymajorgrids=true,
    unbounded coords=jump,
    filter discard warning=false,
    grid style={dashed,gray!35},
]

\nextgroupplot[
    title={GLM4 9B},
    legend style={at={(0.03,0.03)},anchor=south west,draw=black,fill=white},
]
\addplot[name path=glmFinalUpper, draw=none, forget plot] table[x=noise_p, y expr={\thisrow{n}>1 ? \thisrow{mean}+\thisrow{sd_plot} : nan}] {\glmfinaldata};
\addplot[name path=glmFinalLower, draw=none, forget plot] table[x=noise_p, y expr={\thisrow{n}>1 ? \thisrow{mean}-\thisrow{sd_plot} : nan}] {\glmfinaldata};
\addplot[blue!35, fill opacity=0.50, forget plot] fill between[of=glmFinalUpper and glmFinalLower];
\addplot[color=blue!75!black, mark=*, mark size=2.2pt, line width=1.8pt, forget plot]
    table[x=noise_p, y=mean] {\glmfinaldata};
\addlegendimage{
    legend image code/.code={
        \draw[blue!75!black, line width=1.8pt] (0cm,0cm) -- (0.5cm,0cm);
        \fill[blue!75!black] (0.25cm,0cm) circle (1.6pt);
    }
}
\addlegendentry{Group Rollout Noise}
\addplot[black, dashed, line width=1.0pt, forget plot] coordinates {(-0.1,\glmBaselineFinal) (0.6,\glmBaselineFinal)};
\addlegendimage{
    legend image code/.code={
        \draw[black, dashed, line width=1.0pt] (0cm,0cm) -- (0.5cm,0cm);
    }
}
\addlegendentry{No Noise (\pgfmathprintnumber[fixed,precision=3]{\glmBaselineFinal})}

\nextgroupplot[
    title={Qwen3 8B},
    legend style={at={(0.03,0.03)},anchor=south west,draw=black,fill=white},
]
\addplot[name path=qwenFinalUpper, draw=none, forget plot] table[x=noise_p, y expr={\thisrow{n}>1 ? \thisrow{mean}+\thisrow{sd_plot} : nan}] {\qwenfinaldata};
\addplot[name path=qwenFinalLower, draw=none, forget plot] table[x=noise_p, y expr={\thisrow{n}>1 ? \thisrow{mean}-\thisrow{sd_plot} : nan}] {\qwenfinaldata};
\addplot[orange!45, fill opacity=0.50, forget plot] fill between[of=qwenFinalUpper and qwenFinalLower];
\addplot[color=orange!90!black, mark=*, mark size=2.2pt, line width=1.8pt, forget plot]
    table[x=noise_p, y=mean] {\qwenfinaldata};
\addlegendimage{
    legend image code/.code={
        \draw[orange!90!black, line width=1.8pt] (0cm,0cm) -- (0.5cm,0cm);
        \fill[orange!90!black] (0.25cm,0cm) circle (1.6pt);
    }
}
\addlegendentry{Group Rollout Noise}
\addplot[black, dashed, line width=1.0pt, forget plot] coordinates {(-0.1,\qwenBaselineFinal) (0.6,\qwenBaselineFinal)};
\addlegendimage{
    legend image code/.code={
        \draw[black, dashed, line width=1.0pt] (0cm,0cm) -- (0.5cm,0cm);
    }
}
\addlegendentry{No Noise (\pgfmathprintnumber[fixed,precision=3]{\qwenBaselineFinal})}

\end{groupplot}
\end{tikzpicture}
    \caption{Validation reward at the common evaluation checkpoint, rollout step 259, for the recovered group-rollout runs. Shaded regions indicate $\pm 1$ sample standard deviation where multiple seeds are available. Qwen3~8B uses the expanded 3--6-seed cohort, including 4 baseline seeds; GLM4~9B uses 3 through $p{=}0.10$ and 1 above.}
    \label{fig:noise-sweep-final}
\end{figure}

\Cref{fig:noise-sweep-final} shows that low to moderate noise can remain close to the clean baseline at a common checkpoint. At rollout step 259, GLM4~9B with group-level entire-rollout noise at $p{=}0.10$ achieves $0.895$, compared to $0.884$ for the clean baseline. Qwen3~8B reaches $0.883$ versus $0.884$ without noise. Averaging the last two common evaluations gives $0.884$ at $10\%$ noise and $0.890$ for the Qwen3~8B baseline (\Cref{tab:fixed-checkpoints}).

These descriptive comparisons do not establish that noise improves generalization or prevents overfitting. The seed intervals concern a fixed set of 90 validation problems, and no untouched test-set or problem-resampling analysis is available. We discuss the information retained by group-rollout corruption in \Cref{sec: why can models learn with noisy rewards}.

\subsection{The effect of noise structure}

\begin{table}[t]
    \centering
    \caption{Noise-type comparison runs on MBPP, evaluated against ground-truth unit tests. Best and final denote the highest and last checkpoint validation reward; for the recovered Qwen3~8B and GLM4~9B baselines and group-rollout runs, both are restricted to evaluation steps through 259, and final is step 259. The Qwen3~8B baseline and group-rollout rows use the expanded seed cohort of \Cref{tab:fixed-checkpoints}. $\pm$ values are standard deviations across seeds when available.}
    \label{tab: overview}
    \vspace{4pt}
    \fontsize{9.5}{11.5}\selectfont
    \renewcommand{\arraystretch}{1.10}
    \setlength{\tabcolsep}{3pt}
    \begin{tabular}{llcccc}
    \toprule
    \textbf{Model} & \textbf{Setup} & \textbf{Seeds} & \textbf{Best} & \textbf{Final} & \textbf{Steps-to-best} \\
    \midrule
    \multicolumn{6}{l}{\textit{Base model (before training)}} \\
    GLM4 9B & Base model & $3$ & $0.609 \pm 0.010$ & $0.609 \pm 0.010$ & $\mathrm{NA}$ \\
    Qwen3 8B & Base model & $3$ & $0.487 \pm 0.011$ & $0.487 \pm 0.011$ & $\mathrm{NA}$ \\
    \midrule
    \multicolumn{6}{l}{\textit{Baselines (ground-truth unit tests)}} \\
    GLM4 9B & Baseline & $3$ & $0.905 \pm 0.002$ & $0.884 \pm 0.006$ & $159 \pm 34.6$ \\
    Qwen3 8B & Baseline & $4$ & $0.900 \pm 0.010$ & $0.884 \pm 0.013$ & $209 \pm 60.0$ \\
    Llama 3.1 8B & Baseline & $2$ & $0.658 \pm 0.001$ & $0.505 \pm 0.025$ & $59 \pm 28.2$ \\
    \midrule
    \multicolumn{6}{l}{\textit{Controlled noise ($p{=}0.10$)}} \\
    \multirow{4}{*}{GLM4 9B} & Group rollout & $3$ & $0.900 \pm 0.006$ & $0.895 \pm 0.012$ & $212 \pm 80.8$ \\
     & Group unit test & $1$ & $0.891$ & $0.869$ & $139$\\
     & Sample rollout & $1$ & $0.866$ & $0.835$ & $59$ \\
     & Unit test & $1$ & $0.875$ & $0.864$ & $119$ \\
    \cline{2-6} \\[-8pt]
    \multirow{4}{*}{Qwen3 8B} & Group rollout & $6$ & $0.891 \pm 0.015$ & $0.883 \pm 0.019$ & $209 \pm 62.6$ \\
    & Group unit test & $2$ & $0.893 \pm 0.003$ & $0.881 \pm 0.007$ & $229 \pm 14.1$ \\
    & Sample rollout & $2$ & $0.864 \pm 0.011$ & $0.863 \pm 0.012$ & $249 \pm 14.1$ \\
    & Unit test & $2$ & $0.854 \pm 0.013$ & $0.852 \pm 0.014$ & $229 \pm 14.1$ \\
    \cline{2-6} \\[-7pt]
    Llama 3.1 8B & Group rollout & $1$ & $0.651$ & $0.542$ & $19$ \\
    \midrule
    \multicolumn{6}{l}{\textit{Model-based verifier}} \\
    \multirow{2}{*}{Qwen3 8B} & Verifier: Qwen3 4B & $1$ & $0.704$ & $0.652$ & $39$ \\
     & Verifier: Qwen3 30B-A3B & $1$ & $0.871$ & $0.869$ & $179$ \\
    \bottomrule
    \end{tabular}
\end{table}

\Cref{tab: overview} shows the final and best validation reward for each noise type. We compare the four controlled noise modes at a fixed noise rate of $p{=}0.10$, a moderate level, and the model-based verifiers, to see how the type of noise impacts the results.

\paragraph{Controlled noise.} For Qwen3~8B and GLM4~9B, group-level noise (modes c and d) gives higher recorded best and final rewards than sample-level noise (modes a and b). Most comparisons use only 1--2 seeds, so the magnitude and reliability of these differences remain uncertain. The Llama experiment covers only group-rollout noise and does not independently compare the four structures.

\paragraph{Model noise.} With the 30B verifier, peak validation reward is $0.871$, compared to $0.900$ for the clean baseline; with the 4B verifier, it is $0.704$. These single-seed comparisons demonstrate different training outcomes for two judges. Their accuracy and precision also differ, but the comparison does not isolate which properties of the judges cause the performance gap (\Cref{sec: precision vs recall}).

We include results for convergence of GLM4~9B and Qwen3~8B in \Cref{app: time to best}, Llama~3.1~8B in \Cref{tab: overview,app: llama-results}, and Qwen3~4B in \Cref{app: model size results}.

\section{Discussion}
We examine the scope and interpretation of these results: their preliminary extension to scientific reasoning, the learning signal retained by group-level corruption, and what the model-based verifier metrics can establish.

\subsection{Do the results hold for other domains?}
For a preliminary check beyond coding, we consider Graduate-Level Google-Proof Q\&A~(GPQA)~\citep{reinGPQAGraduateLevelGoogleProof2023}, which contains graduate-level scientific reasoning problems with multiple-choice answers. We do not include mathematics; prior work reports varied noise effects for Qwen models in that domain~\citep{shaoSpuriousRewardsRethinking2025,zhuNoisyDataDestructive2026}, and rule-based equivalence checks can reject valid answers~\citep{xuTinyVReducingFalse2025}.

\ifdefined\ishandshake
\begin{wraptable}{R}{0.34\textwidth}
\else
\begin{wraptable}{r}{0.34\textwidth}
\fi
    \centering
    \caption{Best validation reward on GPQA Diamond for Qwen3~8B trained with GSPO under group rollout noise (one seed per setting). Both noisy settings are close to the clean baseline.}
    \label{tab:gpqa-results}
    \vspace{4pt}
    \renewcommand{\arraystretch}{1.12}
    \setlength{\tabcolsep}{8pt}
    \begin{tabular}{lr}
        \toprule
        \textbf{Setup} & \textbf{Best} \\
        \midrule
        Base model & $0.540$ \\
        No noise & $0.600$ \\
        Noise $p{=}0.05$ & $0.604$ \\
        Noise $p{=}0.30$ & $0.603$ \\
        \bottomrule
    \end{tabular}
\end{wraptable}

\paragraph{Experimental setup.} We remove the diamond problems (198 problems) from the main subset (448 problems before, 250 problems after). We then use the 250 remaining problems for training and the 198 diamond problems for validation. We train Qwen3~8B with group rollout noise at $p{=}0.05$ and $p{=}0.30$ along with a noise-free baseline. GPQA uses an 8192-token training response limit, compared with 4096 for MBPP, and Group Sequence Policy Optimization~(GSPO)~\citep{zhengGroupSequencePolicy2025} instead of GRPO, with the aim of stabilizing training (\Cref{app: background}). We do not compare the optimizers on this task.

\paragraph{Results.}
\Cref{tab:gpqa-results} reports peak validation reward: $0.540$ for the base model, $0.600$ after clean training, $0.604$ at $p{=}0.05$, and $0.603$ at $p{=}0.30$. The small differences between trained models are descriptive single-seed observations subject to checkpoint selection. They do not establish a benefit from noise or greater robustness of binary rewards than unit-test rewards.

\paragraph{Observation.} These results suggest that imperfect verification can also support post-training in scientific reasoning. One model, two noise settings, and the change from GRPO to GSPO are insufficient to establish a common robustness threshold across domains.

\subsection{Why can models learn with noisy rewards?}\label{sec: why can models learn with noisy rewards}
Whole-group corruption has a simple relationship to the clean advantages. We derive it below and separate this property from the hypothesis that noise improves optimization or generalization.

\paragraph{Controlled noise.}
For a fixed sampled group without format penalties, a full flip changes rewards from $r_i$ to $1-r_i$, exactly negating its normalized advantage vector $\mathbf A$. Thus $\mathbf A'$ equals $\mathbf A$ with probability $1-p$ and $-\mathbf A$ with probability $p$, giving $\mathbb E[\mathbf A'\mid\mathbf r]=(1-2p)\mathbf A$. The expected unclipped policy-gradient contribution therefore remains aligned with the clean contribution for $p<0.5$. This explains why useful learning signals can survive substantial group-level corruption, without implying a benefit from adding noise. \Cref{app:group-noise-analysis} gives the full distributions with retained format penalties and shows where clipping invalidates simple gradient scaling.

All four symmetric modes have the same conditional expected test reward, $\mathbb E[r_i'\mid\mathbf M]=p+(1-2p)r_i$, for a test-scored response. Their differences arise from correlations between corrupted outcomes and from normalizing each realized group. Since normalization is nonlinear, equal expected rewards do not imply equal expected advantages. Sample-level corruption can alter within-group rankings without preserving or reversing the entire advantage vector. This distinction provides an algebraic reason to analyze noise structure, but does not by itself prove which structure gives better generalization. Moreover, the unchanged $-0.25$ format penalty preserves information about producing a code block even at $p=0.5$. Since early responses often fail to finish within the token budget, the measured gains cannot be fully separated into formatting and coding improvements with the available aggregate records.

\paragraph{Equal error rates can produce different learning signals.}
For a fixed group without format penalties, the asymmetric implementation draws independent shared decisions with $a=\mathrm{FNR}$ and $b=\mathrm{FPR}$. Its normalized advantage vector satisfies
\[
\mathbf A'=\begin{cases}
    \mathbf A, & \text{with probability }(1-a)(1-b),\\
    -\mathbf A, & \text{with probability }ab,\\
    \mathbf 0, & \text{with probability }a+b-2ab.
\end{cases}
\]
Exactly one decision makes all rewards identical, removing every within-group contrast. Consequently, exchanging $a$ and $b$ leaves this distribution unchanged. When $a=b=p$, the mean is $(1-2p)\mathbf A$, as in the main sweep, but the conditional covariance is half as large. Thus identical marginal error rates can produce different distributions of learning signals: group comparisons may be preserved, reversed, or erased. Verifier accuracy alone does not describe these differences. Retained penalties and clipping require the fuller analysis in \Cref{app:group-noise-analysis}.

\paragraph{The Ackley function.}
We include a two-dimensional illustration of optimization with noisy rewards. A Gaussian policy with fixed sampling variance is optimized on the Ackley function using a vanilla policy-gradient objective with group-normalized advantages (\Cref{app: ackley function optimization}). The recorded trajectories run for 500 steps from three starting points on a circle around the global minimum.

In \Cref{fig:ackley-trajectories}, the noiseless trajectories remain near their starting basins, while some noisy trajectories travel toward the global minimum. This experiment uses additive reward noise, rather than the whole-group inversions of our main sweep. It has no held-out objective and therefore does not test generalization, prevention of overfitting, or the proposed mechanism in LLM training. The trajectories are an illustration, not a replicated estimate of optimization benefit.

\begin{figure}[t]
    \centering
    \includegraphics[width=\linewidth]{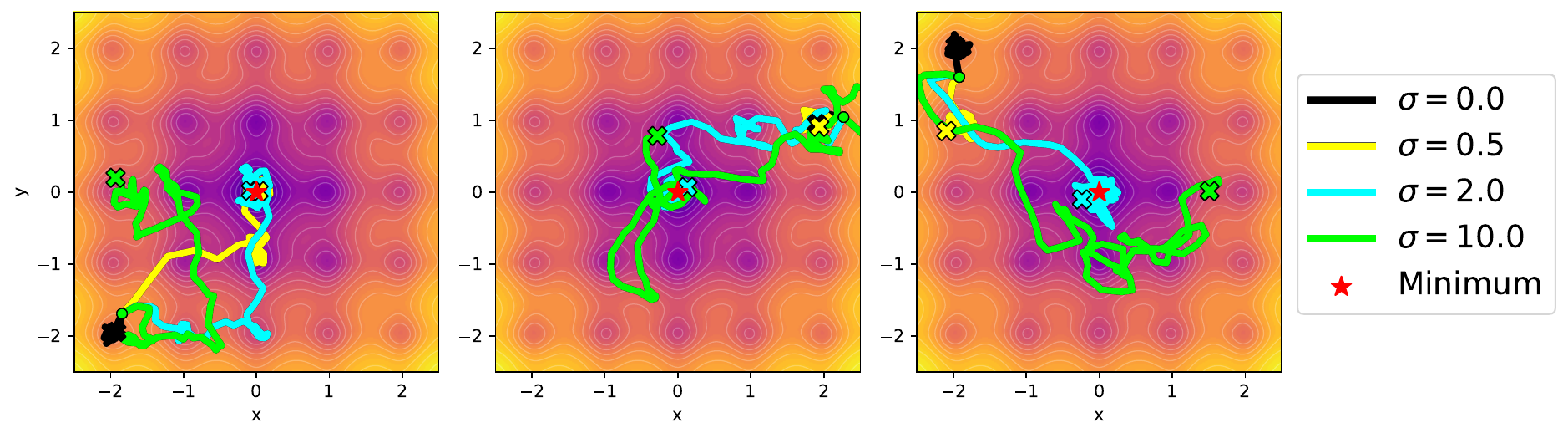}
    \caption{Recorded Ackley optimization trajectories from three starting points under additive reward noise. Here $\sigma$ denotes the plotted reward-noise scale. Some noisy trajectories leave the starting basin; these examples do not test whole-group corruption or generalization.}
    \label{fig:ackley-trajectories}
\end{figure}

\paragraph{Observation.} Resampled group corruption can retain useful advantage information. Whether noise also improves generalization through regularization remains untested; related optimization results are discussed in \Cref{app: benefits of noise and flat minima}.

\subsection{What do the verifier metrics establish?}
\label{sec: precision vs recall}
\begin{figure}[t]
    \centering
    \begin{tikzpicture}
\def\evalLW{1.8pt}
\def\rolloutLW{1.5pt}
\pgfplotsset{table/search path={.,shared/diagram/judge}}

\pgfplotstableread[col sep=comma]{arxiv-shared--diagram--judge--judge_4b_eval_clean.csv}\fourBeval
\pgfplotstableread[col sep=comma]{arxiv-shared--diagram--judge--judge_30b_eval_clean.csv}\thirtyBeval
\pgfplotstableread[col sep=comma]{arxiv-shared--diagram--judge--judge_4b_rollout_smooth.csv}\fourBrollout
\pgfplotstableread[col sep=comma]{arxiv-shared--diagram--judge--judge_30b_rollout_smooth.csv}\thirtyBrollout

\begin{groupplot}[
    group style={
        group size=3 by 1,
        horizontal sep=1.4cm,
    },
    scale only axis,
    width=0.25\columnwidth,
    height=0.34\columnwidth,
    xmin=0, xmax=260,
    xlabel={Steps},
    xtick={0,50,100,150,200,250},
    ymajorgrids=true,
    grid style={dashed,gray!35},
    tick label style={font=\footnotesize},
]

\nextgroupplot[
    title={Reward},
    ymin=0.50, ymax=1.0,
    ytick={0.50,0.60,0.70,0.80,0.90,1.00},
    yticklabels={0.50,0.60,0.70,0.80,0.90,1.00},
]
\addplot[color=blue!75!black, mark=none, line width=\evalLW, forget plot]
    table[x=step, y=mbpp_unittests] {\fourBeval};
\addplot[color=orange!90!black, mark=none, line width=\evalLW, forget plot]
    table[x=step, y=mbpp_unittests] {\thirtyBeval};
\addplot[color=blue!75!black, mark=none, line width=\rolloutLW, dashed, opacity=0.5, forget plot]
    table[x=step, y=raw_reward] {\fourBrollout};
\addplot[color=orange!90!black, mark=none, line width=\rolloutLW, dashed, opacity=0.5, forget plot]
    table[x=step, y=raw_reward] {\thirtyBrollout};
\addplot[black, dashdotted, line width=\rolloutLW, forget plot]
    coordinates {(0,0.900) (260,0.900)};
\addplot[color=blue!75!black, only marks, mark=*, mark size=2pt, forget plot]
    coordinates {(39,0.704)};
\node[blue!75!black, font=\small, anchor=south west, xshift=1pt, yshift=-3.0pt] at (axis cs:39,0.704) {0.704};
\addplot[color=orange!90!black, only marks, mark=*, mark size=2pt, forget plot]
    coordinates {(179,0.871)};
\node[orange!90!black, font=\small, below=2pt] at (axis cs:179,0.871) {0.871};

\nextgroupplot[
    title={Precision},
    ymin=0.5, ymax=1.0,
    ytick={0.50,0.60,0.70,0.80,0.90,1.00},
    yticklabels={0.50,0.60,0.70,0.80,0.90,1.00},
]
\addplot[color=blue!75!black, mark=none, line width=\evalLW, forget plot]
    table[x=step, y=precision] {\fourBeval};
\addplot[color=orange!90!black, mark=none, line width=\evalLW, forget plot]
    table[x=step, y=precision] {\thirtyBeval};
\addplot[color=blue!75!black, mark=none, line width=\rolloutLW, dashed, opacity=0.5, forget plot]
    table[x=step, y=precision] {\fourBrollout};
\addplot[color=orange!90!black, mark=none, line width=\rolloutLW, dashed, opacity=0.5, forget plot]
    table[x=step, y=precision] {\thirtyBrollout};

\nextgroupplot[
    title={Recall},
    ymin=0.90, ymax=1.00,
    ytick={0.90,0.92,0.94,0.96,0.98,1.00},
    yticklabels={0.90,0.92,0.94,0.96,0.98,1.00},
    legend to name=judgelegmain,
    legend columns=3,
    legend style={
        cells={anchor=west},
        draw=none,
        fill=none,
        font=\small,
        row sep=1pt,
        /tikz/every even column/.append style={column sep=10pt},
    },
]
\addplot[color=blue!75!black, mark=none, line width=\evalLW, forget plot]
    table[x=step, y=recall] {\fourBeval};
\addplot[color=orange!90!black, mark=none, line width=\evalLW, forget plot]
    table[x=step, y=recall] {\thirtyBeval};
\addplot[color=blue!75!black, mark=none, line width=\rolloutLW, dashed, opacity=0.5, forget plot]
    table[x=step, y=recall] {\fourBrollout};
\addplot[color=orange!90!black, mark=none, line width=\rolloutLW, dashed, opacity=0.5, forget plot]
    table[x=step, y=recall] {\thirtyBrollout};
\addlegendimage{
    legend image code/.code={
        \draw[blue!75!black, line width=\evalLW] (0cm,0cm) -- (0.5cm,0cm);
    }
}
\addlegendentry{Qwen3 4B}
\addlegendimage{
    legend image code/.code={
        \draw[orange!90!black, line width=\evalLW] (0cm,0cm) -- (0.5cm,0cm);
    }
}
\addlegendentry{Qwen3 30B}
\addlegendimage{
    legend image code/.code={
        \draw[black, dashdotted, line width=\rolloutLW] (0cm,0cm) -- (0.5cm,0cm);
    }
}
\addlegendentry{No noise (0.900)}
\addlegendimage{
    legend image code/.code={
        \draw[gray, line width=\rolloutLW] (0cm,0cm) -- (0.5cm,0cm);
    }
}
\addlegendentry{Eval}
\addlegendimage{
    legend image code/.code={
        \draw[gray, dashed, line width=\rolloutLW] (0cm,0cm) -- (0.5cm,0cm);
    }
}
\addlegendentry{Rollout}
\addlegendimage{
    legend image code/.code={
        \fill[gray] (0.25cm,0cm) circle (1.8pt);
    }
}
\addlegendentry{Best val. reward}

\end{groupplot}
\node[anchor=north, inner sep=0pt] at ([yshift=-1.25cm]group c2r1.south) {\pgfplotslegendfromname{judgelegmain}};
\end{tikzpicture}
    \caption{Training Qwen3~8B with model-based verifier rewards on MBPP (one seed per judge). \textbf{Left:} validation reward against executed unit tests (solid) and rollout reward from the judge (dashed). Peak validation rewards are 0.871 with the 30B judge and 0.704 with the 4B judge, versus 0.900 for the clean baseline. \textbf{Center and right:} verifier precision and recall on each judge's corresponding policy outputs; these are different response distributions.}
    \label{fig:judge-plots-main}
\end{figure}

\Cref{fig:judge-plots-main} shows the two model-based verifier runs on MBPP; we did not run analogous judge experiments on GPQA. The 4B run peaks early and subsequently declines, whereas the 30B run reaches a final reward of $0.869$, compared with the clean baseline's $0.884$. Both judges have recall above $90\%$ on their respective policy outputs, while the 4B judge has lower precision. Because precision depends on the prevalence of correct outputs, it can change as a consequence of policy quality. Without evaluating both judges on shared response pools, these curves do not isolate intrinsic judge quality from the policies they train.

The rising judge reward alongside declining clean validation reward in the 4B run is consistent with reward overoptimization~\citep{gaoScalingLawsReward2023}. However, the curves compare training prompts scored by a judge with validation prompts scored by execution. They do not establish a particular exploitation mechanism or show that false negatives encourage useful exploration.

\citet{xuTinyVReducingFalse2025} find benefits from reducing false negatives in mathematical verification. Our experiments do not contradict that result: the task and verifier interventions differ, and neither judge comparison identifies a general precision--recall trade-off (\Cref{app: tinyv discussion}).

\paragraph{Observation.} High observed recall alone does not ensure effective training. The asymmetric sweep (\Cref{tab:rebuttal-asymmetric}) also cannot isolate the relative effects of FPR and FNR: swapping them leaves the advantage distribution unchanged for groups without format penalties. \Cref{app:group-noise-analysis} treats mixed-format groups and the normalization stabilizer explicitly.

\section{Conclusion}

We studied reward corruption in coding RLVR, with a preliminary scientific-reasoning check. In the expanded Qwen3~8B group-rollout sweep, observed mean reward over two fixed late evaluations differs from the clean baseline by less than $1$ percentage point at the tested rates through $20\%$, and by about $2$ points at $30\%$. Confidence intervals allow larger losses. One model-based verifier run reaches a peak reward of $0.871$, compared to $0.900$ for clean training. These observations show that imperfect verification can support improvement in the tested settings. Our analysis identifies the advantage information preserved by whole-group corruption and the restrictions on interpreting asymmetric group noise. These findings motivate assessing verifiers through both aggregate error rates and the reward contrasts they induce within a rollout group, while measuring downstream quality against a separate reference.

\paragraph{Limitations.} Most replication concerns one model and one resampled corruption process. MBPP has only 90 validation problems; seed intervals do not quantify generalization to new problems, and peak selection can be optimistic. GPQA changes the optimizer and uses one seed per setting. Pass@k records and exact per-run source revisions are incomplete. Retained format penalties confound the separation of formatting and correctness gains. We do not establish a verifier-accuracy threshold, robustness to persistent biases, a general precision--recall preference, or improved generalization from noise.

\clearpage

\printbibliography

\clearpage
\appendix
\crefalias{section}{appendix}
\crefalias{subsection}{appendix}
\crefalias{subsubsection}{appendix}

\section{Reproducibility}
Qwen3 before our RL training performs worse under our evaluation protocol than reported in the Qwen team's technical report \citep{yangQwen3TechnicalReport2025}. This cross-protocol difference does not itself establish an evaluation error in either study. \ifdefined\isiclr The supplementary archive provides the recorded evaluation settings and reward implementations.\else We provide a script that reproduces our numbers with \texttt{tinker}.\footnote{\url{https://gist.github.com/AndreasPlesner-hs/1b66fb59091e62496906b0b35960fdc3}}\fi\ Our evaluation applies the model's chat template \citep{huggingface-chat-templates}; the reported comparisons use this prompting protocol.

\paragraph{Run records and selection.} The accompanying data include individual seed scores, run manifests and configuration records, and scripts for reproducing the tables from recorded W\&B histories. All 37 expanded Qwen3~8B peak scores match runs in the original and later projects. The clean baseline uses seeds 1, 3, 11 and 12; $5\%$ uses seeds 1--3; $10\%$ uses seeds 1--3 and 10--12; the remaining settings use one original seed and seeds 10--12. We preserve this recorded cohort, with additional and repeated run candidates documented separately; this is a retrospective cohort, not a prospectively specified inclusion rule. All 37 runs have evaluations at steps 239 and 259. Our retrospective fixed-step analysis averages these two scores within each run, then reports the mean and sample standard deviation across seeds, without selecting a checkpoint by validation performance. The archives do not identify exact per-run source revisions; recovered source snapshots and their limitations are documented with the analysis scripts. Reconstructing aggregate results from these records does not establish that every historical training run can be reproduced exactly.

\paragraph{Evaluation uncertainty.} All MBPP evaluations use the same 90 validation problems. The reported standard deviations and confidence intervals describe variation across training seeds on this fixed problem set; they do not include uncertainty from sampling new problems. The 16 generations per problem reduce generation-sampling uncertainty but are not independent problems. No untouched test-set evaluation or prespecified noninferiority margin is available, so small observed mean differences do not establish equivalence within a chosen practical tolerance.

\section{Training details}
\label{sec:training-details}

\subsection{Final hyperparameters}

\Cref{tab:hyperparameters} lists the hyperparameters used across all experiments unless otherwise noted.

\begin{table}[ht]
    \centering
    \caption{Hyperparameters used for all experiments. MBPP experiments use GRPO; GPQA experiments use GSPO.}
    \label{tab:hyperparameters}
    \vspace{4pt}
    \begin{tabular}{lc}
    \toprule
    Hyperparameter & Value \\
    \midrule
    \multicolumn{2}{l}{\textit{Optimization}} \\
    Optimizer & Adam \\
    Learning rate & $1 \times 10^{-6}$ \\
    LR schedule & Constant \\
    Weight decay & $0.10$ \\
    $\beta_1$, $\beta_2$ & $0.90$, $0.98$ \\
    \midrule
    \multicolumn{2}{l}{\textit{GRPO / GSPO}} \\
    Clip range ($\epsilon_{\text{low}}$, $\epsilon_{\text{high}}$) & $0.20$, $0.28$ \\
    KL coefficient & $0.00$ \\
    Entropy coefficient & $0.00$ \\
    Rollouts per prompt & $16$ \\
    Prompts per rollout batch (recovered MBPP runs) & $48$ \\
    Global batch size & $96$ \\
    \midrule
    \multicolumn{2}{l}{\textit{Sampling (training)}} \\
    Temperature & $1.00$ \\
    Top-$p$ & $1.00$ \\
    Max response length & $4096$ (MBPP) / $8192$ (GPQA) \\
    \midrule
    \multicolumn{2}{l}{\textit{Evaluation}} \\
    Samples per prompt & $16$ \\
    Temperature & $0.70$ \\
    Top-$p$ & $1.00$ \\
    Max response length & $8192$ \\
    Eval interval & Every $10$ or $20$ rollout steps \\
    \bottomrule
    \end{tabular}
\end{table}

The original MBPP runs and later clean baselines evaluate every 20 rollout steps; most later noisy runs evaluate every 10. The curves use the shared index grid 0, 19, 39, and so on through 259. Runs were configured for 261 or 500 rollout iterations; the expanded peak table retains all recorded maxima, including later evaluations through step 479 where available. Matching rollout steps controls the nominal sampling budget, not exact generated-token counts or wall-clock time. The recovered GLM checkpoint is \texttt{GLM-Z1-9B-0414}, referred to as GLM4~9B in the paper.

\subsection{Hyperparameter sensitivity}

\Cref{tab:hp-sensitivity} compares training performance under alternative sampling hyperparameters for GLM4~9B and Qwen3~8B. The main setting (top-$p{=}1.00$, temperature${=}1.00$) is reproduced from \Cref{tab: overview} for reference.

In the recorded runs, reducing top-$p$ from $1.00$ to $0.95$ lowers GLM4~9B's best reward from $0.905$ to $0.817$ and final reward from $0.884$ to $0.764$. Qwen3~8B's best reward falls from $0.900$ to $0.829$. Lowering the temperature from $1.00$ to $0.70$ also lowers both models' observed scores, particularly Qwen3~8B's ($0.803$ vs.\ $0.900$ best). Reduced sampling diversity is one possible explanation, but these runs do not isolate the mechanism. Increasing the global batch size from 96 to 768 yields lower observed peak scores for both models (GLM4~9B: $0.888$ vs.\ $0.905$; Qwen3~8B: $0.845$ vs.\ $0.900$); the rounded best and final scores coincide in each larger-batch run. The small seed counts, including single runs for temperature and batch-size changes, limit conclusions about stability and convergence speed.

\begin{table}[ht]
    \centering
    \caption{Sensitivity to sampling hyperparameters for baseline (no noise) training on MBPP. $T$ denotes temperature, $B$ the global batch size. All runs use 16 rollouts per prompt. The main setting rows are reproduced from \Cref{tab: overview}.}
    \label{tab:hp-sensitivity}
    \vspace{4pt}
    \begin{tabular}{llccc}
    \toprule
    Model & Hyperparameters & Seeds & Best & Final \\
    \midrule
    \multicolumn{5}{l}{\textit{GLM4 9B}} \\
    & top-$p{=}1.00$, $T{=}1.00$, $B{=}96$ (main) & $3$ & $0.905 \pm 0.002$ & $0.884 \pm 0.006$ \\
    & top-$p{=}0.95$, $T{=}1.00$, $B{=}96$ & $2$ & $0.817 \pm 0.005$ & $0.764 \pm 0.002$ \\
    & top-$p{=}1.00$, $T{=}0.70$, $B{=}96$ & $1$ & $0.895$ & $0.881$ \\
    & top-$p{=}1.00$, $T{=}1.00$, $B{=}768$ & $1$ & $0.888$ & $0.888$ \\
    \midrule
    \multicolumn{5}{l}{\textit{Qwen3 8B}} \\
    & top-$p{=}1.00$, $T{=}1.00$, $B{=}96$ (main) & $4$ & $0.900 \pm 0.010$ & $0.884 \pm 0.013$ \\
    & top-$p{=}0.95$, $T{=}1.00$, $B{=}96$ & $2$ & $0.829 \pm 0.025$ & $0.827 \pm 0.022$ \\
    & top-$p{=}1.00$, $T{=}0.70$, $B{=}96$ & $1$ & $0.803$ & $0.769$ \\
    & top-$p{=}1.00$, $T{=}1.00$, $B{=}768$ & $1$ & $0.845$ & $0.845$ \\
    \bottomrule
    \end{tabular}
\end{table}

\section{Additional evaluation results}
\label{app:additional-results}

\paragraph{Expanded peak results.} \Cref{tab:rebuttal-peak} reports the additional Qwen3~8B group-rollout runs on MBPP. Each seed contributes its highest recorded validation reward. These maxima include different training durations and evaluation frequencies, giving unequal opportunities for checkpoint selection. We therefore report a separate comparison at fixed evaluation steps in \Cref{app: time to best}. Means, sample standard deviations, seed counts and $p$-values are reproduced from the recorded seed scores; the statistical test is the two-sided independent equal-variance $t$-test used in Excel (\texttt{T.TEST(...,2,2)}). The comparisons share a baseline and are unadjusted for multiple testing. They characterize the observed peak scores rather than establish equivalent performance or a universal noise threshold.

\begin{table}[htbp]
    \centering\small
    \caption{Expanded peak mean unit-test rewards for Qwen3~8B on MBPP with group-rollout noise. $n$ is the number of seeds; $\Delta$ is the difference from the clean baseline in percentage points, with a Welch 95\% confidence interval. Each run contributes its best recorded score, including later checkpoints where available.}
    \label{tab:rebuttal-peak}
    \vspace{4pt}
\begin{tabular}{crrrcr}
\toprule
Noise rate & $n$ & Peak score (mean $\pm$ SD) & $\Delta$ (pp) & 95\% CI (pp) & $p$-value \\
\midrule
0 & 4 & $0.9003 \pm 0.0104$ & --- & --- & --- \\
0.05 & 3 & $0.8879 \pm 0.0220$ & -1.24 & $[-5.9, +3.5]$ & 0.3601 \\
0.10 & 6 & $0.8939 \pm 0.0168$ & -0.63 & $[-2.6, +1.4]$ & 0.5246 \\
0.15 & 4 & $0.8922 \pm 0.0102$ & -0.81 & $[-2.6, +1.0]$ & 0.3085 \\
0.20 & 4 & $0.8960 \pm 0.0071$ & -0.42 & $[-2.0, +1.2]$ & 0.5269 \\
0.25 & 4 & $0.8837 \pm 0.0051$ & -1.66 & $[-3.2, -0.1]$ & 0.0284 \\
0.30 & 4 & $0.8798 \pm 0.0086$ & -2.05 & $[-3.7, -0.4]$ & 0.0231 \\
0.40 & 4 & $0.8168 \pm 0.0348$ & -8.34 & $[-13.7, -3.0]$ & 0.0037 \\
0.50 & 4 & $0.5988 \pm 0.1132$ & -30.14 & $[-48.1, -12.2]$ & 0.0018 \\
\bottomrule
\end{tabular}
 \end{table}

\paragraph{Full-program correctness.} \Cref{tab:rebuttal-passk} reports the recorded best pass@k over evaluated checkpoints, using 16 generations per validation problem and requiring all three tests to pass. For $c$ successful generations, the standard estimator is $1-\binom{16-c}{k}/\binom{16}{k}$, averaged across problems. The available summaries cover fewer seeds than \Cref{tab:rebuttal-peak}. These are peak metrics and remain subject to checkpoint selection; individual seed, checkpoint and generation records were not recovered, so a common-checkpoint pass@k comparison is unavailable. The aggregate table can be reproduced from the archived summary, but the contributing runs, selected checkpoints and per-problem estimates cannot be independently reconstructed. These results therefore provide descriptive supporting evidence, with weaker provenance than the expanded unit-test reward analysis.

\begin{table}[htbp]
    \centering\small
    \caption{Peak full-program pass@k for Qwen3~8B on MBPP with group-rollout noise. Values are mean $\pm$ standard deviation across $n$ seeds. An SD displayed as $0.000$ is rounded. The metric-specific seed cohort is smaller than that of \Cref{tab:rebuttal-peak}.}
    \label{tab:rebuttal-passk}
    \vspace{4pt}
    \setlength{\tabcolsep}{4pt}
\begin{tabular}{crcccc}
\toprule
Noise rate & $n$ & pass@1 & pass@4 & pass@8 & pass@16 \\
\midrule
0 & 2 & $0.814 \pm 0.005$ & $0.860 \pm 0.017$ & $0.875 \pm 0.018$ & $0.894 \pm 0.008$ \\
0.10 & 2 & $0.841 \pm 0.001$ & $0.891 \pm 0.004$ & $0.907 \pm 0.006$ & $0.922 \pm 0.000$ \\
0.15 & 3 & $0.816 \pm 0.012$ & $0.876 \pm 0.011$ & $0.895 \pm 0.012$ & $0.915 \pm 0.006$ \\
0.20 & 2 & $0.827 \pm 0.001$ & $0.884 \pm 0.001$ & $0.911 \pm 0.003$ & $0.933 \pm 0.000$ \\
0.25 & 2 & $0.818 \pm 0.004$ & $0.868 \pm 0.000$ & $0.882 \pm 0.001$ & $0.894 \pm 0.008$ \\
0.30 & 2 & $0.810 \pm 0.005$ & $0.872 \pm 0.013$ & $0.898 \pm 0.013$ & $0.922 \pm 0.016$ \\
0.40 & 2 & $0.756 \pm 0.012$ & $0.849 \pm 0.005$ & $0.876 \pm 0.000$ & $0.894 \pm 0.008$ \\
0.50 & 2 & $0.551 \pm 0.104$ & $0.715 \pm 0.025$ & $0.744 \pm 0.013$ & $0.767 \pm 0.000$ \\
\bottomrule
\end{tabular}
 \end{table}

\paragraph{Asymmetric noise.} \Cref{tab:rebuttal-asymmetric} reports all 25 settings of the recovered Qwen3~8B sweep. This implementation draws independent group-level decisions $Z_{\mathrm{FN}}\sim\operatorname{Bernoulli}(\mathrm{FNR})$ and $Z_{\mathrm{FP}}\sim\operatorname{Bernoulli}(\mathrm{FPR})$, applying $M'_{i,j}=M_{i,j}(1-Z_{\mathrm{FN}})+(1-M_{i,j})Z_{\mathrm{FP}}$ to the original test outcomes. It differs from the single-decision corruption of the main symmetric design, even when FPR and FNR are equal; format penalties remain unchanged.

The grid is exploratory: all settings use seed 12 and provide no across-seed uncertainty. More fundamentally, without retained format penalties, let $a=\mathrm{FNR}$, $b=\mathrm{FPR}$ and $\hat{\mathbf A}$ denote the clean advantage vector. The corrupted vector equals $\hat{\mathbf A}$, $-\hat{\mathbf A}$ or $\mathbf 0$ with probabilities $(1-a)(1-b)$, $ab$ and $a(1-b)+(1-a)b$, respectively (\Cref{app:group-noise-analysis}). Its full conditional distribution is invariant to exchanging $a$ and $b$, so this design cannot identify different effects of false positives and false negatives on correctness learning in the absence of penalties. Retaining the uniform negative format penalty replaces the zero-advantage states of mixed groups with a formatting signal. Those two states have the same normalized advantage direction, and the swap invariance remains exact when the normalization stabilizer is zero; with a positive stabilizer, their magnitudes can differ. Thus, the penalty does not resolve the intended identification problem. Individual trajectories also vary with sampled groups and noise decisions. Moreover, exchanging rates does not hold verifier accuracy fixed when the pass/fail prevalence differs from one half. The observed differences between swapped settings therefore do not establish a general precision--recall preference.

\begin{table}[htbp]
    \centering\small
    \caption{Peak mean unit-test rewards in the asymmetric-noise sweep. Rows are false-negative rates and columns are false-positive rates. Runs use seed 12 and evaluate every 10 rollout steps through 259, except $\dagger$, interrupted after step 229. $\ddagger$ averages peaks from two records sharing seed 12, as in the original summary.}
    \label{tab:rebuttal-asymmetric}
    \vspace{4pt}
\begin{tabular}{crrrrr}
\toprule
FNR $\backslash$ FPR & 0.05 & 0.15 & 0.30 & 0.40 & 0.50 \\
\midrule
0.05 & 0.892 & 0.906 & 0.905 & 0.887 & 0.878 \\
0.15 & 0.900 & 0.899 & 0.893 & 0.896 & 0.886 \\
0.30 & 0.900 & 0.893 & 0.869 & 0.875 & 0.864 \\
0.40 & 0.891 & 0.901 & 0.895 & 0.851 & 0.823$^{\dagger}$ \\
0.50 & 0.896 & 0.860 & 0.888 & 0.843 & 0.781$^{\ddagger}$ \\
\bottomrule
\end{tabular}
 \end{table}

\section{Prompts}
\label{sec:prompts}

We list the prompts used in our experiments. For MBPP and GPQA, the task description is wrapped using the model's chat template (\texttt{apply\_chat\_template} \citep{huggingface-chat-templates}). For the model-based verifier experiments, we use the following system and user prompts.

\subsection{Model-based verifier: unit test evaluation}

Below we show the single-test version of the judge prompt. The archived configurations do not identify whether requests used single-test or batched judging.

The archived judge models are \texttt{Qwen3-4B} and \texttt{Qwen3-30B-A3B-}\allowbreak\texttt{Instruct-2507}, with temperature 0. The requested response is a JSON object containing a boolean \texttt{status}, as shown below. Recovered source validates the JSON, attempts to extract an embedded JSON object if needed, and assigns a false verdict if parsing still fails; request exceptions return reward 0. The configurations specify up to five attempts with a 5000-second timeout per attempt, but the exact active retry implementation is not archived per run.

Historical judge training code also differs in its reward conventions: an earlier version assigns $0$ to missing code and floors formatted-code rewards at $0.5/T$, whereas a later version uses the $-0.25$ format penalty without that floor. The missing per-run source revisions prevent resolving this detail for each judge run. The controlled-noise derivation in \Cref{app:group-noise-analysis} concerns the synthetic corruption rules and should not be assumed to describe these judge training rewards. Clean validation uses the executed-unit-test metric described in the main text.

\begin{quote}
\small
\textbf{System:} You are an expert code evaluator evaluating the quality of a model's response against a test case.
Your task is to determine whether the GENERATED CODE satisfies the TEST CASE. I.e., whether the assert statement would pass.
Answer with a JSON object with the key ``status'' and a boolean value.

Instructions:
\begin{itemize}
\item Carefully read the TEST CASE to understand what is required.
\item Evaluate the GENERATED CODE against this TEST CASE only (ignore anything outside it).
\end{itemize}

\textbf{User:}\\
GENERATED CODE:\\
\texttt{\{generated\_code\}}\\[0.5em]
TEST CASE:\\
\texttt{assert \{function\_name\}(\{test\_input\}) == \{test\_output\}}
\end{quote}

\subsection{GPQA: answer extraction}

For GPQA, the model is presented with the question and multiple-choice options via the chat template. The reward is computed by extracting the selected option letter from the model's response using a series of regex patterns and comparing it against the correct answer. No model-based verifier is used for GPQA evaluation.

\section{Background}
\label{app: background}

In RLVR, a language model is trained using reinforcement learning, where a verifier determines the reward for each generated response---for example, executing unit tests for code or checking a multiple-choice answer. This section describes the training objective.

\paragraph{GRPO.}

We use GRPO as our training objective. Unlike the earlier Proximal Policy Optimization (PPO) \citep{schulmanProximalPolicyOptimization2017}, GRPO omits the value network and instead estimates advantages by comparing rollouts within a group sampled for the same prompt.

Concretely, for each prompt $x$, GRPO samples a group of $G$ responses $\{y_1, \ldots, y_G\} \sim \pi_{\theta_{\text{old}}}(\cdot \mid x)$ and computes a reward $r_i = R(x, y_i)$ for each response. The advantage for response $y_i$ is computed by normalizing rewards within the group:
\begin{equation}
    \hat{A}_i = \frac{r_i - \mu(\mathbf{r})}{\sigma(\mathbf{r}) + \epsilon}\label{eq: advantage estimate}
\end{equation}
where $\mu(\mathbf{r})$ and $\sigma(\mathbf{r})$ are the mean and standard deviation of the group rewards, and $\epsilon$ is a small constant for numerical stability.

The clipped GRPO objective is
\begin{equation}
    \begin{aligned}
    J(\theta)&=\mathbb{E}\left[\frac{1}{G}\sum_{i=1}^{G}\frac{1}{|y_i|}\sum_{t=1}^{|y_i|}\ell_{i,t}(\theta)\right],\\
    \ell_{i,t}(\theta)&=\min\!\left(\rho_{i,t}\hat A_i,
    \operatorname{clip}(\rho_{i,t},1-\epsilon_{\mathrm{low}},1+\epsilon_{\mathrm{high}})\hat A_i\right),
    \end{aligned}
    \label{eq:grpo-objective}
\end{equation}
where the expectation is over training prompts and responses sampled from $\pi_{\theta_{\mathrm{old}}}$, and $\rho_{i,t}=\pi_\theta(y_{i,t}\mid x,y_{i,<t})/\pi_{\theta_{\mathrm{old}}}(y_{i,t}\mid x,y_{i,<t})$. The recovered configurations use per-response token averaging, clipping bounds $\epsilon_{\mathrm{low}}=0.20$ and $\epsilon_{\mathrm{high}}=0.28$, and zero KL and entropy coefficients.

\paragraph{GSPO.} For the scientific reasoning experiments, we use Group Sequence Policy Optimization (GSPO) \citep{zhengGroupSequencePolicy2025} instead of GRPO with the aim of stabilizing training. GSPO retains group-normalized advantages but uses sequence-level importance ratios and clipping. We do not compare the optimizers on this task. We refer the reader to \citet{zhengGroupSequencePolicy2025} for technical details.

\section{Conditional advantage distributions under group-level noise}
\label{app:group-noise-analysis}

This section analyzes the two group-rollout corruption rules used in the controlled experiments. All distributions and expectations below condition on a fixed prompt, its sampled responses and clean test outcomes; the only randomness is the corruption decision. When discussing gradients, we also fix the current and sampling policies and treat advantages as detached, as in GRPO. These conditional identities do not assume that different training runs visit the same policies or responses.

\subsection{Centering and retained format penalties}

Assume $G\ge2$. Let $m_i\in\{0,1\}$ indicate whether response $i$ has a nonnegative, test-scored reward, and let $q_i\in[0,1]$ be that reward when $m_i=1$. Execution failures have $q_i=0$; missing-code responses have $m_i=0$ and retain the penalty $c=-1/4$. The value assigned to $q_i$ when $m_i=0$ is immaterial. Write
\begin{equation}
    \mathbf r=\mathbf m\odot\mathbf q+c(\mathbf 1-\mathbf m),\qquad
    H=I-\frac{1}{G}\mathbf 1\mathbf 1^{\mathsf T},\qquad
    \mathbf z=H\mathbf r,\quad \mathbf h=H\mathbf m.
\end{equation}
For a centered vector $\mathbf v$, define
\begin{equation}
    s_d(\mathbf v)=\sqrt{\frac{\|\mathbf v\|_2^2}{d}},\qquad
    F(\mathbf v)=\frac{\mathbf v}{s_d(\mathbf v)+\eta},\qquad
    \mathbf A=F(\mathbf z),
    \label{eq:noise-normalization}
\end{equation}
where $\eta>0$ is the normalization stabilizer, distinct from the clipping bounds. The identities hold for either population standard deviation ($d=G$) or sample standard deviation ($d=G-1$). The recovered normalization source uses the sample convention and $\eta=10^{-6}$; exact source revisions are unavailable per run. In the idealized $\eta=0$ case, define $F(\mathbf0)=\mathbf0$.

Suppose corruption maps every eligible reward $q_i$ to $\alpha q_i+\beta$ and leaves $c$ unchanged. Centering gives the general identity
\begin{equation}
    H\mathbf r'=\alpha\mathbf z+
    [\beta+c(\alpha-1)]\mathbf h,
    \qquad
    \mathbf A'=F\!\left(\alpha\mathbf z+
    [\beta+c(\alpha-1)]\mathbf h\right).
    \label{eq:noise-mask-transform}
\end{equation}
Each realized reward vector must be normalized separately: in general, $\mathbb E[\mathbf A']\ne F(H\mathbb E[\mathbf r'])$.

\subsection{Symmetric corruption with one shared decision}

In the main sweep, $Z\sim\operatorname{Bernoulli}(p)$ and eligible rewards become $q_i'=(1-Z)q_i+Z(1-q_i)$. Define the flipped advantage vector
\begin{equation}
    \mathbf A_{\mathrm F}=F(-\mathbf z+(1-2c)\mathbf h)
    =F(-\mathbf z+\tfrac32\mathbf h).
\end{equation}
The exact conditional law and moments, including penalties, are
\begin{align}
    \mathbf A'&=\begin{cases}
        \mathbf A,&\text{with probability }1-p,\\
        \mathbf A_{\mathrm F},&\text{with probability }p,
    \end{cases} \label{eq:symmetric-advantage-law}\\
    \mathbb E[\mathbf A']&=(1-p)\mathbf A+p\mathbf A_{\mathrm F},\\
    \operatorname{Cov}(\mathbf A')&=p(1-p)
    (\mathbf A_{\mathrm F}-\mathbf A)
    (\mathbf A_{\mathrm F}-\mathbf A)^{\mathsf T}.
\end{align}
If all responses are test-scored, $\mathbf h=\mathbf0$ and $F(-\mathbf z)=-F(\mathbf z)$, including for $\eta>0$. Thus
\begin{equation}
    \mathbf A'=(1-2Z)\mathbf A,\qquad
    \mathbb E[\mathbf A']=(1-2p)\mathbf A,\qquad
    \operatorname{Cov}(\mathbf A')=4p(1-p)\mathbf A\mathbf A^{\mathsf T}.
    \label{eq:symmetric-no-format}
\end{equation}
The corruption preserves the expected advantage direction for $p<1/2$ and applies a single shared sign multiplier within a fixed group. This is a property of this particular corruption rule, not a general property of erroneous verifiers.

With mixed test-scored and penalized responses, full-group corruption does not generally negate the advantages. Even at $p=1/2$, eligible rewards have conditional mean $1/2$, whereas penalized rewards remain $-1/4$. A format-related signal also survives normalization: if $k=\sum_i m_i$ satisfies $0<k<G$, and $\bar q$ is the mean eligible score, then
\begin{equation}
    \mathbf h^{\mathsf T}\mathbf z
       =k(1-k/G)(\bar q-c)>0,\qquad
    \mathbf h^{\mathsf T}(-\mathbf z+(1-2c)\mathbf h)
       =k(1-k/G)(1-\bar q-c)>0.
\end{equation}
Both realizations therefore favor test-scored responses over penalized responses on average within the group. If all responses are penalized, every advantage is zero. Consequently, $50\%$ corruption does not remove all reward information from the implemented training setup.

\subsection{Asymmetric corruption with two shared decisions}

Let $a=\mathrm{FNR}$ and $b=\mathrm{FPR}$. The exploratory implementation draws independent $U\sim\operatorname{Bernoulli}(a)$ and $V\sim\operatorname{Bernoulli}(b)$, shared across the group, and applies both decisions to the original outcomes. Eligible rewards become
\begin{equation}
    q_i'=(1-U-V)q_i+V.
\end{equation}
Applying \Cref{eq:noise-mask-transform} yields the full conditional distribution:
\begin{equation}
\begin{array}{ccll}
    (U,V)&\text{Probability}&H\mathbf r'&\mathbf A'\\[2pt]
    (0,0)&(1-a)(1-b)&\mathbf z&\mathbf A\\
    (1,0)&a(1-b)&-c\mathbf h&\mathbf A_0:=F(-c\mathbf h)\\
    (0,1)&(1-a)b&(1-c)\mathbf h&\mathbf A_1:=F((1-c)\mathbf h)\\
    (1,1)&ab&-\mathbf z+(1-2c)\mathbf h&\mathbf A_{\mathrm F}.
\end{array}
\label{eq:asymmetric-advantage-law}
\end{equation}
Its mean is the probability-weighted sum of these four vectors; its covariance is $\sum_j w_j\mathbf A_j\mathbf A_j^{\mathsf T}-\boldsymbol\mu\boldsymbol\mu^{\mathsf T}$, where $w_j$ are the listed probabilities and $\boldsymbol\mu$ is their weighted mean.

Without penalties, $\mathbf h=\mathbf0$, so exactly one decision makes every reward identical and all advantages zero. In this case
\begin{equation}
    \mathbf A'=\begin{cases}
        \mathbf A,&\text{with probability }(1-a)(1-b),\\
        -\mathbf A,&\text{with probability }ab,\\
        \mathbf0,&\text{with probability }a+b-2ab,
    \end{cases}
    \label{eq:asymmetric-no-format}
\end{equation}
and
\begin{equation}
    \mathbb E[\mathbf A']=(1-a-b)\mathbf A,\qquad
    \operatorname{Cov}(\mathbf A')=
    [a(1-a)+b(1-b)]\mathbf A\mathbf A^{\mathsf T}.
\end{equation}
The entire distribution is invariant to exchanging $a$ and $b$. When $a=b=p$, its mean matches the one-decision rule, but it has zero-advantage probability $2p(1-p)$ and inversion probability $p^2$, and half the conditional covariance. Equal marginal error rates therefore do not make the two interventions equivalent.

Retained penalties do not resolve this identifiability limitation. In a mixed group, let $s=s_d(\mathbf h)>0$. For $c=-1/4$,
\begin{equation}
    \mathbf A_0=\frac{\mathbf h}{s+4\eta},\qquad
    \mathbf A_1=\frac{\mathbf h}{s+\tfrac45\eta}.
\end{equation}
These vectors have the same direction and coincide exactly when $\eta=0$, in which case the entire distribution remains invariant under exchanging $a$ and $b$, even with penalties. For $\eta>0$, swapping the rates only swaps the probabilities of these two nearby vectors. In particular,
\begin{align}
    \mathbb E_{a,b}[\mathbf A']-\mathbb E_{b,a}[\mathbf A']
       &=(a-b)(\mathbf A_0-\mathbf A_1),\\
    \mathbf A_0-\mathbf A_1
       &=-\frac{\tfrac{16}{5}\eta\,\mathbf h}
       {(s+4\eta)(s+\tfrac45\eta)}.
    \label{eq:asymmetric-swap-difference}
\end{align}
For sample normalization, $s^2=k(G-k)/[G(G-1)]$, so with $G=16$ every mixed group has $s\ge1/4$. At the recovered stabilizer $\eta=10^{-6}$, the relative difference $\|\mathbf A_0-\mathbf A_1\|_2/\|\mathbf A_1\|_2$ is at most $1.28\times10^{-5}$ in exact arithmetic. This is an algebraic bound, not a measured effect. Individual training trajectories can still differ, but swapped-rate runs under this design do not isolate a substantive preference between false positives and false negatives in correctness learning.

\subsection{Unclipped gradients, clipping, and optimizer updates}

For a fixed group and current policy, the unclipped surrogate gradient is linear in the advantages:
\begin{equation}
    g_\theta(\mathbf A)=\frac1G\sum_{i=1}^G
       \frac{A_i}{|y_i|}\sum_{t=1}^{|y_i|}\nabla_\theta\rho_{i,t}.
\end{equation}
At $\theta=\theta_{\mathrm{old}}$, each ratio is one and $\nabla_\theta\rho_{i,t}=\nabla_\theta\log\pi_\theta(y_{i,t}\mid x,y_{i,<t})$. Without penalties, \Cref{eq:symmetric-no-format,eq:asymmetric-no-format} give
\begin{equation}
    \mathbb E[g_\theta(\mathbf A')]
    =\begin{cases}
        (1-2p)g_\theta(\mathbf A),&\text{one decision},\\
        (1-a-b)g_\theta(\mathbf A),&\text{two decisions}.
    \end{cases}
    \label{eq:noise-gradient-scaling}
\end{equation}
With penalties, linearity still gives $\mathbb E[g_\theta(\mathbf A')]=g_\theta(\mathbb E[\mathbf A'])$, using the full mixtures above, rather than a universal scalar attenuation.

Clipping limits the sign-reversal interpretation. Let $L=1-\epsilon_{\mathrm{low}}$, $U=1+\epsilon_{\mathrm{high}}$, and $[v]_+=\max(v,0)$. For one token, the clipped surrogate in \Cref{eq:grpo-objective} satisfies
\begin{align}
    \ell(\rho,A)
       &=\rho A-[A]_+[\rho-U]_+-[-A]_+[L-\rho]_+,\\
    \ell(\rho,A)+\ell(\rho,-A)
       &=-|A|D(\rho),\qquad
       D(\rho)=[\rho-U]_+ + [L-\rho]_+.
    \label{eq:noise-clipping-identity}
\end{align}
Thus, for groups without penalties, the exact conditional expected clipped surrogates are
\begin{equation}
    \mathbb E[\ell(\rho,A')]=\begin{cases}
       (1-2p)\ell(\rho,A)-p|A|D(\rho),&\text{one decision},\\
       (1-a-b)\ell(\rho,A)-ab|A|D(\rho),&\text{two decisions}.
    \end{cases}
    \label{eq:noise-clipped-expectation}
\end{equation}
These identities apply token by token and hence to the full response-averaged objective. Within the clipping interval, $D(\rho)=0$ and unclipped scaling holds; at $\theta=\theta_{\mathrm{old}}$ all ratios lie strictly inside this interval. Outside it, an additional term penalizes distance from the interval. In particular, at $p=1/2$ the expected one-decision surrogate is $-|A|D(\rho)/2$: its gradient is zero inside the interval but can be nonzero outside, pointing toward the interval as a function of $\rho$. Statements about derivatives exclude the clipping boundaries, where the objective is not differentiable. With retained penalties the exact expected clipped objective is instead the weighted sum of $\ell(\rho,A_j)$ over \Cref{eq:symmetric-advantage-law,eq:asymmetric-advantage-law}; substituting only the mean advantage is generally invalid.

These are conditional statements about raw surrogate gradients at a fixed policy. Reusing a corrupted batch makes later policies depend on its realized corruption. Adam additionally maps gradients and their history through nonlinear moment normalization, and weight decay contributes a separate update. Neither expected-gradient attenuation nor a single-group sign reversal therefore implies a correspondingly scaled or reversed parameter update, identical training dynamics, escape from local minima, or improved generalization. The preserved expected direction provides a direct explanation for why this specific resampled corruption can permit learning below $50\%$ noise; it does not establish a practical verifier-accuracy threshold or a regularization benefit.

\section{In-depth discussion}

\subsection{Noise and flat minima: possible connections}
\label{app: benefits of noise and flat minima}
As discussed in \Cref{sec: why can models learn with noisy rewards,app:group-noise-analysis}, group-wide reward complementation reverses normalized advantages when every response receives a test-based reward. Below $50\%$ corruption, the expected unclipped contribution retains the clean direction. This algebraic property explains why the noise can preserve a useful learning signal without requiring a regularization or basin-escape mechanism.

Prior work studies connections between stochastic optimization, noise structure and flatness \citep{keskarLargeBatchTrainingDeep2017,leBayesianPerspectiveGeneralization2018,zhuAnisotropicNoiseStochastic2019}, as well as deliberate perturbations and smoothed objectives \citep{foretSharpnessawareMinimizationEfficiently2020,chaudhariEntropySGDBiasingGradient2017}. Other analyses examine escape from local minima or saddle points under particular assumptions \citep{kleinbergAlternativeViewWhen2018,jinHowEscapeSaddle2017}. These results motivate hypotheses about our training dynamics, but their interventions and assumptions differ from randomly complementing rewards. We do not measure flatness, curvature, basin escape or improved generalization in the trained language models. The Ackley illustration uses additive reward noise and has no training/test distinction; it does not test the group-complementation mechanism or show that it prevents overfitting.

\subsection{The results of \texorpdfstring{\citet{xuTinyVReducingFalse2025}}{Xu et al. (2025)}}
\label{app: tinyv discussion}

\citet{xuTinyVReducingFalse2025} study false negatives in mathematical verification and introduce a lightweight model-based verifier that rechecks responses rejected by a rule-based verifier. Their reported improvements show the value of this intervention in their setting. They do not conflict with our observation that a coding verifier can have high recall while providing poor training rewards.

These are different comparisons: TinyV augments a verifier, whereas our judge experiment changes the model used for every judgment and evaluates each judge on its own trained policy's outputs. Our shared-decision asymmetric experiment also cannot isolate a general precision--recall trade-off (\Cref{app:group-noise-analysis}). Consequently, our results neither refute the value of reducing false negatives nor establish that false positives are intrinsically more harmful. Comparing those effects would require comparable response distributions and an intervention that separates them.

\section{Detailed metrics for the model-based verifiers}
\label{app: complete judge metrics}
\begin{figure}[ht]
    \centering
    \begin{tikzpicture}
\def\evalLW{1.8pt}
\def\rolloutLW{1.5pt}

\pgfplotsset{table/search path={.,shared/diagram/judge}}

\pgfplotstableread[col sep=comma]{arxiv-shared--diagram--judge--judge_4b_eval_clean.csv}\fourBeval
\pgfplotstableread[col sep=comma]{arxiv-shared--diagram--judge--judge_30b_eval_clean.csv}\thirtyBeval
\pgfplotstableread[col sep=comma]{arxiv-shared--diagram--judge--judge_4b_rollout_smooth.csv}\fourBrollout
\pgfplotstableread[col sep=comma]{arxiv-shared--diagram--judge--judge_30b_rollout_smooth.csv}\thirtyBrollout

\begin{groupplot}[
    group style={
        group size=3 by 2,
        horizontal sep=1.6cm,
        vertical sep=0.5cm,
        xlabels at=edge bottom,
        xticklabels at=edge bottom,
    },
    scale only axis,
    width=0.3\columnwidth,
    height=0.20\columnwidth,
    xmin=0, xmax=260,
    xlabel={Steps},
    xtick={0,50,100,150,200,250},
    ymajorgrids=true,
    grid style={dashed,gray!35},
    tick label style={font=\footnotesize},
]

\nextgroupplot[
    ylabel={Reward},
    ymin=0.45, ymax=1.02,
    ytick={0.50,0.60,0.70,0.80,0.90,1.00},
    yticklabels={0.50,0.60,0.70,0.80,0.90,1.00},
]
\addplot[color=blue!75!black, mark=none, line width=\evalLW, forget plot]
    table[x=step, y=mbpp_unittests] {\fourBeval};
\addplot[color=orange!90!black, mark=none, line width=\evalLW, forget plot]
    table[x=step, y=mbpp_unittests] {\thirtyBeval};
\addplot[color=blue!75!black, mark=none, line width=\rolloutLW, dashed, opacity=0.5, forget plot]
    table[x=step, y=raw_reward] {\fourBrollout};
\addplot[color=orange!90!black, mark=none, line width=\rolloutLW, dashed, opacity=0.5, forget plot]
    table[x=step, y=raw_reward] {\thirtyBrollout};
\addplot[black, dashed, line width=\rolloutLW, forget plot]
    coordinates {(0,0.900) (260,0.900)};
\addplot[color=blue!75!black, only marks, mark=*, mark size=2.5pt, forget plot]
    coordinates {(39,0.704)};
\node[blue!75!black, font=\small, anchor=south west, xshift=1pt, yshift=-3.0pt] at (axis cs:39,0.704) {0.704};
\addplot[color=orange!90!black, only marks, mark=*, mark size=2.5pt, forget plot]
    coordinates {(179,0.871)};
\node[orange!90!black, font=\small, below=2pt] at (axis cs:179,0.871) {0.871};

\nextgroupplot[
    ylabel={Accuracy},
    ymin=0.5, ymax=0.96,
    ytick={0.50,0.60,0.70,0.80,0.90},
    yticklabels={0.50,0.60,0.70,0.80,0.90},
]
\addplot[color=blue!75!black, mark=none, line width=\evalLW, forget plot]
    table[x=step, y=accuracy] {\fourBeval};
\addplot[color=orange!90!black, mark=none, line width=\evalLW, forget plot]
    table[x=step, y=accuracy] {\thirtyBeval};
\addplot[color=blue!75!black, mark=none, line width=\rolloutLW, dashed, opacity=0.5, forget plot]
    table[x=step, y=accuracy] {\fourBrollout};
\addplot[color=orange!90!black, mark=none, line width=\rolloutLW, dashed, opacity=0.5, forget plot]
    table[x=step, y=accuracy] {\thirtyBrollout};

\nextgroupplot[
    ylabel={F1},
    ymin=0.65, ymax=1.0,
    ytick={0.70,0.80,0.90,1.00},
    yticklabels={0.70,0.80,0.90,1.00},
]
\addplot[color=blue!75!black, mark=none, line width=\evalLW, forget plot]
    table[x=step, y=f1] {\fourBeval};
\addplot[color=orange!90!black, mark=none, line width=\evalLW, forget plot]
    table[x=step, y=f1] {\thirtyBeval};
\addplot[color=blue!75!black, mark=none, line width=\rolloutLW, dashed, opacity=0.5, forget plot]
    table[x=step, y=f1] {\fourBrollout};
\addplot[color=orange!90!black, mark=none, line width=\rolloutLW, dashed, opacity=0.5, forget plot]
    table[x=step, y=f1] {\thirtyBrollout};

\nextgroupplot[
    ylabel={Precision},
    ymin=0.5, ymax=1.0,
    ytick={0.50,0.60,0.70,0.80,0.90,1.00},
    yticklabels={0.50,0.60,0.70,0.80,0.90,1.00},
]
\addplot[color=blue!75!black, mark=none, line width=\evalLW, forget plot]
    table[x=step, y=precision] {\fourBeval};
\addplot[color=orange!90!black, mark=none, line width=\evalLW, forget plot]
    table[x=step, y=precision] {\thirtyBeval};
\addplot[color=blue!75!black, mark=none, line width=\rolloutLW, dashed, opacity=0.5, forget plot]
    table[x=step, y=precision] {\fourBrollout};
\addplot[color=orange!90!black, mark=none, line width=\rolloutLW, dashed, opacity=0.5, forget plot]
    table[x=step, y=precision] {\thirtyBrollout};

\nextgroupplot[
    ylabel={Recall},
    ymin=0.90, ymax=1.00,
    ytick={0.90,0.92,0.94,0.96,0.98,1.00},
    yticklabels={0.90,0.92,0.94,0.96,0.98,1.00},
    legend to name=judgeleg,
    legend columns=1,
    legend style={
        cells={anchor=west},
        draw=none,
        fill=none,
    },
]
\addplot[color=blue!75!black, mark=none, line width=\evalLW, forget plot]
    table[x=step, y=recall] {\fourBeval};
\addplot[color=orange!90!black, mark=none, line width=\evalLW, forget plot]
    table[x=step, y=recall] {\thirtyBeval};
\addplot[color=blue!75!black, mark=none, line width=\rolloutLW, dashed, opacity=0.5, forget plot]
    table[x=step, y=recall] {\fourBrollout};
\addplot[color=orange!90!black, mark=none, line width=\rolloutLW, dashed, opacity=0.5, forget plot]
    table[x=step, y=recall] {\thirtyBrollout};
\addlegendimage{
    legend image code/.code={
        \draw[blue!75!black, line width=\evalLW] (0cm,0cm) -- (0.5cm,0cm);
    }
}
\addlegendentry{Qwen3 4B (eval)}
\addlegendimage{
    legend image code/.code={
        \draw[orange!90!black, line width=\evalLW] (0cm,0cm) -- (0.5cm,0cm);
    }
}
\addlegendentry{Qwen3 30B (eval)}
\addlegendimage{
    legend image code/.code={
        \draw[blue!75!black, line width=\rolloutLW, dashed, opacity=0.5] (0cm,0cm) -- (0.5cm,0cm);
    }
}
\addlegendentry{Qwen3 4B (rollout)}
\addlegendimage{
    legend image code/.code={
        \draw[orange!90!black, line width=\rolloutLW, dashed, opacity=0.5] (0cm,0cm) -- (0.5cm,0cm);
    }
}
\addlegendentry{Qwen3 30B (rollout)}
\addlegendimage{
    legend image code/.code={
        \draw[black, dashed, line width=\rolloutLW] (0cm,0cm) -- (0.5cm,0cm);
    }
}
\addlegendentry{No noise (0.900)}
\addlegendimage{
    legend image code/.code={
        \fill[black] (0.25cm,0cm) circle (1.8pt);
    }
}
\addlegendentry{Best validation reward}

\nextgroupplot[group/empty plot, ymin=0, ymax=1]

\end{groupplot}
\node at (group c3r1.center |- group c1r2.center) {\pgfplotslegendfromname{judgeleg}};
\end{tikzpicture}
    \caption{Training Qwen3~8B with model-based verifier rewards on MBPP. Solid lines show evaluation metrics; dashed lines show exponentially smoothed rollout metrics. \textbf{Top row, left to right:} validation reward (against ground-truth unit tests) with rollout reward from the model-based verifier, verifier accuracy, and verifier F1 score. \textbf{Bottom row:} verifier precision and recall. The Qwen3~4B verifier (blue) shows lower precision than the 30B verifier (orange) on their respective policy outputs, despite high recall. Their trained policies reach peak validation rewards of 0.704 and 0.871, respectively. Each verifier has one training run; this association does not isolate an effect of precision.}
    \label{fig:judge-plots-complete}
\end{figure}

We show in \Cref{fig:judge-plots-complete} the reward for training and validation when using the model-based verifiers; these overlap with the results in \Cref{sec: precision vs recall}. For training, the rewards are given by the model-based verifier, while the validation rewards come from running the actual unit tests. The accuracy, F1 score, precision, and recall are for the model-based verifier when comparing the pass/fail decisions with running the unit tests.

As discussed in \Cref{sec: precision vs recall}, the 4B verifier has lower observed precision, accuracy and F1 score, with slightly higher recall. High recall alone does not ensure useful training rewards in these runs, but this comparison does not isolate the relative effects of false positives and false negatives. Each judge is measured on outputs from the policy trained with that judge. Precision depends on the prevalence of correct responses, so the measured precision difference can reflect both verifier behavior and differences between the trained policies. The 4B verifier's F1 score across evaluation checkpoints is approximately $0.72$--$0.85$; F1 values from other tasks, such as those of \citet{heAdvancedIFRubricBasedBenchmarking2025}, are not directly comparable because the data and error distributions differ.

\begin{figure}[t]
    \centering
    \begin{tikzpicture}
\begin{axis}[
    ybar=0pt,
    scale only axis,
    width=0.85\columnwidth,
    height=0.425\columnwidth,
    bar width=14pt,
    ymin=0.00, ymax=1.00,
    ylabel={Best reward},
    xmin=-0.55, xmax=1.85,
    xtick={0,1.3},
    xticklabels={Qwen3 4B,Qwen3 8B},
    xticklabel style={font=\small},
    ylabel style={font=\small},
    tick label style={font=\footnotesize},
    ytick={0.00,0.20,0.40,0.60,0.80,1.00},
    yticklabels={0.00,0.20,0.40,0.60,0.80,1.00},
    ymajorgrids=true,
    grid style={dashed,gray!35},
    legend style={
        at={(1.02,0.5)},
        anchor=west,
        legend columns=1,
        draw=none,
        font=\footnotesize
    },
    area legend,
    nodes near coords,
    every node near coord/.append style={
        font=\scriptsize,
        text=black,
        rotate=0,
        anchor=south,
        yshift=3pt
    },
    point meta=y,
    nodes near coords={\pgfmathprintnumber[fixed,fixed zerofill,precision=3]{\pgfplotspointmeta}},
]

\path[fill=blue!4, draw=none] (axis cs:-0.55,1.00) rectangle (axis cs:0.65,0.00);
\path[fill=orange!4, draw=none] (axis cs:0.65,1.00) rectangle (axis cs:2.00,0.00);

\addplot+[fill=gray!45, draw=gray!70!black] coordinates {
    (-0.24, 0.59337)
    (1.06, 0.487)
};

\addplot+[fill=green!55!black, draw=green!35!black] coordinates {
    (-0.12, 0.868725)
    (1.18, 0.900521)
};

\addplot+[fill=blue!60, draw=blue!80!black] coordinates {
    (0.00, 0.8667)
    (1.30, 0.8879)
};

\addplot+[fill=orange!85, draw=orange!60!black] coordinates {
    (0.12, 0.8197)
    (1.42, 0.8713)
};

\addplot+[fill=red!70, draw=red!75!black] coordinates {
    (0.24, 0.8035)
    (1.54, 0.7688)
};

\legend{
    Base model,
    No noise,
    Noise $p{=}0.05$,
    Noise $p{=}0.30$,
    Noise $p{=}0.40$
}
\end{axis}
\end{tikzpicture}
    \caption{Best validation reward for Qwen3~4B and 8B under group rollout noise in the original model-size comparison. This figure retains the original 8B cohort, rather than the expanded cohort in \Cref{tab:rebuttal-peak}; the ordering at $p{=}0.40$ changes with that expansion. The comparison does not establish a model-capacity effect.}
    \label{fig:model-size}
\end{figure}

\section{Convergence rate}\label{app: time to best}
\Cref{fig:convergence-curves} compares the training curves of the clean baseline and group rollout noise at $p{=}0.1$. We extend this comparison across noise levels in \Cref{fig:convergence-all-noise} and report fixed-step summaries in \Cref{tab:fixed-checkpoints}, using all 37 expanded Qwen3~8B runs. The mean over evaluation steps 239 and 259 is $0.890$ without noise and $0.884$ at $p{=}0.10$. It falls to $0.870$ at $p{=}0.30$, $0.806$ at $p{=}0.40$, and $0.355$ at $p{=}0.50$, with substantial variation across seeds at the highest rate. Thus the small observed mean differences at moderate noise also occur without selecting each run's peak. The intervals still permit larger losses: at $p{=}0.20$, the late-training difference is approximately $-0.6$ percentage points with a 95\% interval of $[-2.9,+1.8]$. These finite-budget curves establish neither noninferiority within one point nor asymptotic convergence; some runs improve after step 259, and the expanded peak results include later observations where available.

\begin{figure}[ht]
    \centering
    \begin{tikzpicture}
\pgfplotsset{table/search path={.,shared/results/training,../../results/training}}
\pgfplotstableread[col sep=comma]{arxiv-shared--results--training--convergence_p010.csv}\curvedata

\begin{groupplot}[
    group style={group size=2 by 1, horizontal sep=1.6cm},
    width=0.525\columnwidth,
    height=0.35\columnwidth,
    xmin=0, xmax=265,
    ymin=0.45, ymax=0.95,
    ytick={0.50,0.60,0.70,0.80,0.90},
    yticklabels={0.50,0.60,0.70,0.80,0.90},
    xlabel={Rollout step},
    ylabel={Validation Reward},
    ymajorgrids=true,
    grid style={dashed,gray!35},
    tick label style={font=\footnotesize},
    label style={font=\small},
]

\nextgroupplot[
    title={GLM4-9B},
    legend style={at={(0.97,0.03)},anchor=south east,font=\footnotesize,draw=black,fill=white},
]
\addplot[name path=glmBlUpper, draw=none, forget plot] table[x=step, y expr=\thisrow{glm4_9b_baseline_mean}+\thisrow{glm4_9b_baseline_std}] {\curvedata};
\addplot[name path=glmBlLower, draw=none, forget plot] table[x=step, y expr=\thisrow{glm4_9b_baseline_mean}-\thisrow{glm4_9b_baseline_std}] {\curvedata};
\addplot[black!30, fill opacity=0.50, forget plot] fill between[of=glmBlUpper and glmBlLower];
\addplot[color=black, mark=square*, mark size=1.6pt, line width=1.8pt, forget plot]
    table[x=step, y=glm4_9b_baseline_mean] {\curvedata};
\addlegendimage{
    legend image code/.code={
        \draw[black, line width=1.8pt] (0cm,0cm) -- (0.5cm,0cm);
        \fill[black] (0.25cm,0cm) +(-1.4pt,-1.4pt) rectangle +(1.4pt,1.4pt);
    }
}
\addlegendentry{No Noise}
\addplot[name path=glmNUpper, draw=none, forget plot] table[x=step, y expr=\thisrow{glm4_9b_noisy_mean}+\thisrow{glm4_9b_noisy_std}] {\curvedata};
\addplot[name path=glmNLower, draw=none, forget plot] table[x=step, y expr=\thisrow{glm4_9b_noisy_mean}-\thisrow{glm4_9b_noisy_std}] {\curvedata};
\addplot[blue!35, fill opacity=0.50, forget plot] fill between[of=glmNUpper and glmNLower];
\addplot[color=blue!75!black, mark=*, mark size=1.8pt, line width=1.8pt, forget plot]
    table[x=step, y=glm4_9b_noisy_mean] {\curvedata};
\addlegendimage{
    legend image code/.code={
        \draw[blue!75!black, line width=1.8pt] (0cm,0cm) -- (0.5cm,0cm);
        \fill[blue!75!black] (0.25cm,0cm) circle (1.6pt);
    }
}
\addlegendentry{Group Rollout $p{=}0.10$}

\nextgroupplot[
    title={Qwen3-8B},
    legend style={at={(0.97,0.03)},anchor=south east,font=\footnotesize,draw=black,fill=white},
]
\addplot[name path=qwenBlUpper, draw=none, forget plot] table[x=step, y expr=\thisrow{qwen3_8b_baseline_mean}+\thisrow{qwen3_8b_baseline_std}] {\curvedata};
\addplot[name path=qwenBlLower, draw=none, forget plot] table[x=step, y expr=\thisrow{qwen3_8b_baseline_mean}-\thisrow{qwen3_8b_baseline_std}] {\curvedata};
\addplot[black!30, fill opacity=0.50, forget plot] fill between[of=qwenBlUpper and qwenBlLower];
\addplot[color=black, mark=square*, mark size=1.6pt, line width=1.8pt, forget plot]
    table[x=step, y=qwen3_8b_baseline_mean] {\curvedata};
\addlegendimage{
    legend image code/.code={
        \draw[black, line width=1.8pt] (0cm,0cm) -- (0.5cm,0cm);
        \fill[black] (0.25cm,0cm) +(-1.4pt,-1.4pt) rectangle +(1.4pt,1.4pt);
    }
}
\addlegendentry{No Noise}
\addplot[name path=qwenNUpper, draw=none, forget plot] table[x=step, y expr=\thisrow{qwen3_8b_noisy_mean}+\thisrow{qwen3_8b_noisy_std}] {\curvedata};
\addplot[name path=qwenNLower, draw=none, forget plot] table[x=step, y expr=\thisrow{qwen3_8b_noisy_mean}-\thisrow{qwen3_8b_noisy_std}] {\curvedata};
\addplot[orange!45, fill opacity=0.50, forget plot] fill between[of=qwenNUpper and qwenNLower];
\addplot[color=orange!90!black, mark=*, mark size=1.8pt, line width=1.8pt, forget plot]
    table[x=step, y=qwen3_8b_noisy_mean] {\curvedata};
\addlegendimage{
    legend image code/.code={
        \draw[orange!90!black, line width=1.8pt] (0cm,0cm) -- (0.5cm,0cm);
        \fill[orange!90!black] (0.25cm,0cm) circle (1.6pt);
    }
}
\addlegendentry{Group Rollout $p{=}0.10$}

\end{groupplot}
\end{tikzpicture}
    \caption{Recovered training curves for group rollout noise at $p{=}0.1$ through the common rollout step 259. Shaded regions indicate $\pm 1$ sample standard deviation across seeds.}
    \label{fig:convergence-curves}
\end{figure}

\begin{figure}[htbp]
    \centering
    \begin{tikzpicture}
\pgfplotsset{table/search path={.,../../results/wandb/rebuttal,shared/results/wandb/rebuttal}}
\pgfplotstableread[col sep=comma]{arxiv-shared--results--wandb--rebuttal--training_curve_summary.csv}\allnoisecurves

\pgfplotstablecreatecol[
    create col/assign/.code={
        \getthisrow{sd}\curvesd
        \ifx\curvesd\empty\def\curvesd{0}\fi
        \pgfkeyslet{/pgfplots/table/create col/next content}\curvesd
    }
]{plot_sd}\allnoisecurves

\newcommand{\drawnoisecurve}[2]{%
    \def\curvehasband{0}%
    \pgfplotstableforeachcolumnelement{noise_probability}\of\allnoisecurves\as\curvelevel{%
        \ifdim\curvelevel pt=#1pt
            \pgfplotstablegetelem{\pgfplotstablerow}{n}\of\allnoisecurves
            \ifnum\pgfplotsretval>1 \gdef\curvehasband{1}\fi
        \fi
    }%
    \ifnum\curvehasband=1
        \addplot[name path=curveupper, draw=none, forget plot]
            table[
                x expr={abs(\thisrow{noise_probability}-#1)<0.00001 ? \thisrow{step} : nan},
                y expr={\thisrow{n}>1 ? \thisrow{mean}+\thisrow{plot_sd} : nan}
            ] {\allnoisecurves};
        \addplot[name path=curvelower, draw=none, forget plot]
            table[
                x expr={abs(\thisrow{noise_probability}-#1)<0.00001 ? \thisrow{step} : nan},
                y expr={\thisrow{n}>1 ? \thisrow{mean}-\thisrow{plot_sd} : nan}
            ] {\allnoisecurves};
        \addplot[fill=#2, fill opacity=0.18, draw=none, forget plot]
            fill between[of=curveupper and curvelower];
    \fi
    \addplot[color=#2, mark=none, line width=1.2pt, forget plot]
        table[
            x expr={abs(\thisrow{noise_probability}-#1)<0.00001 ? \thisrow{step} : nan},
            y=mean
        ] {\allnoisecurves};
}

\begin{groupplot}[
    group style={
        group size=4 by 2,
        horizontal sep=0.40cm,
        vertical sep=0.90cm,
        x descriptions at=edge bottom,
        y descriptions at=edge left,
    },
    scale only axis,
    width=0.205\columnwidth,
    height=0.19\columnwidth,
    xmin=0, xmax=259,
    ymin=-0.10, ymax=0.95,
    xtick={0,100,200},
    ytick={0.0,0.3,0.6,0.9},
    yticklabels={0.0,0.3,0.6,0.9},
    xlabel={Rollout step},
    ylabel={Validation reward},
    ymajorgrids=true,
    grid style={dashed,gray!35},
    tick label style={font=\footnotesize},
    label style={font=\small},
    title style={font=\small},
    unbounded coords=jump,
    filter discard warning=false,
]

\nextgroupplot[title={$p=0.05$}]
\drawnoisecurve{0}{black}
\drawnoisecurve{0.05}{orange!90!black}
\nextgroupplot[title={$p=0.10$}]
\drawnoisecurve{0}{black}
\drawnoisecurve{0.10}{orange!90!black}
\nextgroupplot[title={$p=0.15$}]
\drawnoisecurve{0}{black}
\drawnoisecurve{0.15}{orange!90!black}
\nextgroupplot[title={$p=0.20$}]
\drawnoisecurve{0}{black}
\drawnoisecurve{0.20}{orange!90!black}
\nextgroupplot[title={$p=0.25$}]
\drawnoisecurve{0}{black}
\drawnoisecurve{0.25}{orange!90!black}
\nextgroupplot[title={$p=0.30$}]
\drawnoisecurve{0}{black}
\drawnoisecurve{0.30}{orange!90!black}
\nextgroupplot[title={$p=0.40$}]
\drawnoisecurve{0}{black}
\drawnoisecurve{0.40}{orange!90!black}
\nextgroupplot[title={$p=0.50$}]
\drawnoisecurve{0}{black}
\drawnoisecurve{0.50}{orange!90!black}
\end{groupplot}

\path (group c1r1.north west) -- (group c4r1.north east)
    node[midway, above=0.60cm, font=\small]
    {\tikz[baseline=-0.5ex]\draw[black,line width=1.2pt] (0,0)--(0.4,0);
     \ No noise\hspace{1.2em}
     \tikz[baseline=-0.5ex]\draw[orange!90!black,line width=1.2pt] (0,0)--(0.4,0);
     \ Group-rollout noise};
\end{tikzpicture}
    \caption{Qwen3~8B validation trajectories for the expanded seed cohort in \Cref{tab:rebuttal-peak}, compared at common rollout steps. Each panel compares one group-rollout noise rate with the clean baseline. Bands show sample standard deviations where at least two seeds are available. Some resumed runs have no early evaluations, so contributing seed counts increase along a curve; all 37 runs have the last two evaluations.}
    \label{fig:convergence-all-noise}
\end{figure}

\begin{table}[htbp]
    \centering\small
    \caption{Qwen3~8B mean unit-test rewards for the expanded cohort at fixed evaluation steps. The late-training metric averages steps 239 and 259 within each seed before aggregation; $\pm$ denotes sample SD across seeds. $\Delta$ late is its difference from the clean baseline with a Welch 95\% confidence interval; at $p{=}0.50$ the normal-theory interval extends beyond the attainable range because of the large variation across seeds. The best column uses available evaluations through 259 and still has unequal checkpoint-selection opportunities.}
    \label{tab:fixed-checkpoints}
    \vspace{4pt}
    \setlength{\tabcolsep}{3pt}
\begin{tabular}{crcccc}
\toprule
Noise rate & $n$ & Best through 259 & At step 259 & Mean of 239, 259 & $\Delta$ late (pp) [95\% CI] \\
\midrule
0 & 4 & $0.900 \pm 0.010$ & $0.884 \pm 0.013$ & $0.890 \pm 0.014$ & --- \\
0.05 & 3 & $0.888 \pm 0.022$ & $0.885 \pm 0.023$ & $0.882 \pm 0.018$ & $-0.8$ $[-4.3, +2.7]$ \\
0.10 & 6 & $0.891 \pm 0.015$ & $0.883 \pm 0.019$ & $0.884 \pm 0.018$ & $-0.6$ $[-3.0, +1.8]$ \\
0.15 & 4 & $0.891 \pm 0.009$ & $0.884 \pm 0.009$ & $0.885 \pm 0.008$ & $-0.5$ $[-2.7, +1.7]$ \\
0.20 & 4 & $0.893 \pm 0.012$ & $0.885 \pm 0.007$ & $0.884 \pm 0.012$ & $-0.6$ $[-2.9, +1.8]$ \\
0.25 & 4 & $0.884 \pm 0.005$ & $0.878 \pm 0.010$ & $0.876 \pm 0.007$ & $-1.4$ $[-3.5, +0.8]$ \\
0.30 & 4 & $0.880 \pm 0.009$ & $0.871 \pm 0.006$ & $0.870 \pm 0.004$ & $-2.0$ $[-4.2, +0.2]$ \\
0.40 & 4 & $0.817 \pm 0.035$ & $0.812 \pm 0.032$ & $0.806 \pm 0.031$ & $-8.4$ $[-13.1, -3.8]$ \\
0.50 & 4 & $0.599 \pm 0.113$ & $0.347 \pm 0.371$ & $0.355 \pm 0.339$ & $-53.5$ $[-107.3, +0.3]$ \\
\bottomrule
\end{tabular}
 \end{table}

\section{Model size}\label{app: model size results}
\Cref{fig:model-size} compares Qwen3~4B and Qwen3~8B under group rollout noise in the original model-size cohort. At $p{=}0.05$, both models' recorded peak scores are close to their clean baselines. The original comparison gives $0.769$ for 8B and $0.804$ for 4B at $p{=}0.40$, but the expanded 8B peak mean is approximately $0.817$ (\Cref{tab:rebuttal-peak}), reversing that ordering. The cohorts and checkpoint-selection opportunities differ, and the 4B results lack a corresponding expanded analysis. These descriptive observations do not support a conclusion that either size is intrinsically more robust to noise.

\section{Response length}
\Cref{fig:response-length} shows the median response length of Qwen3~8B and GLM4~9B during training with clean rewards and with group rollout noise at $p{=}0.1$. Both conditions show decreasing median lengths with qualitatively similar trajectories. This comparison does not establish that noise has no effect on response length or on the rate of change.

\paragraph{Why does the response length decrease over time?} One of the key observations in the DeepSeek R1 paper was that the model tends to generate longer responses during training, which they link to the model's ability to solve harder problems more accurately \citep[Figure 1 (b)]{deepseek-aiDeepSeekR1IncentivizingReasoning2025}. Specifically, the authors write the following.
\begin{quote}
    As shown in Figure 1(b), DeepSeek-R1-Zero exhibits a steady increase in thinking time throughout training, driven solely by intrinsic adaptation rather than external modifications. Leveraging long CoT, the model progressively refines its reasoning, generating hundreds to thousands of tokens to explore and improve its problem-solving strategies.

    The increase in thinking time fosters the autonomous development of sophisticated behaviors. Specifically, DeepSeek-R1-Zero increasingly exhibits advanced reasoning strategies such as reflective reasoning and systematic exploration of alternative solutions. \citep[p.~5]{deepseek-aiDeepSeekR1IncentivizingReasoning2025}
\end{quote}

\begin{figure}[t]
    \centering
    \begin{tikzpicture}
\pgfplotsset{table/search path={.,shared/diagram/convergence}}
\pgfplotstableread[col sep=comma]{arxiv-shared--diagram--convergence--data--response_length_group_rollout_p0.1.csv}\curvedata

\begin{groupplot}[
    group style={group size=2 by 1, horizontal sep=2.1cm},
    width=0.525\columnwidth,
    height=0.35\columnwidth,
    xmin=0, xmax=265,
    ymin=500, ymax=2900,
    ytick={1000,2000},
    yticklabels={1k, 2k},
    xlabel={Step},
    ylabel={Median Response Length},
    ymajorgrids=true,
    grid style={dashed,gray!35},
    tick label style={font=\footnotesize},
    label style={font=\small},
]

\nextgroupplot[
    title={GLM4-9B},
    legend style={at={(0.97,0.97)},anchor=north east,font=\footnotesize,draw=black,fill=white},
]
\addplot[name path=glmBlUpper, draw=none, forget plot] table[x=step, y expr=\thisrow{glm4_9b_baseline_mean}+\thisrow{glm4_9b_baseline_std}] {\curvedata};
\addplot[name path=glmBlLower, draw=none, forget plot] table[x=step, y expr=\thisrow{glm4_9b_baseline_mean}-\thisrow{glm4_9b_baseline_std}] {\curvedata};
\addplot[black!30, fill opacity=0.50, forget plot] fill between[of=glmBlUpper and glmBlLower];
\addplot[color=black, mark=square*, mark size=1.6pt, line width=1.8pt, forget plot]
    table[x=step, y=glm4_9b_baseline_mean] {\curvedata};
\addlegendimage{
    legend image code/.code={
        \draw[black, line width=1.8pt] (0cm,0cm) -- (0.5cm,0cm);
        \fill[black] (0.25cm,0cm) +(-1.4pt,-1.4pt) rectangle +(1.4pt,1.4pt);
    }
}
\addlegendentry{No Noise}
\addplot[name path=glmNUpper, draw=none, forget plot] table[x=step, y expr=\thisrow{glm4_9b_noisy_mean}+\thisrow{glm4_9b_noisy_std}] {\curvedata};
\addplot[name path=glmNLower, draw=none, forget plot] table[x=step, y expr=\thisrow{glm4_9b_noisy_mean}-\thisrow{glm4_9b_noisy_std}] {\curvedata};
\addplot[blue!35, fill opacity=0.50, forget plot] fill between[of=glmNUpper and glmNLower];
\addplot[color=blue!75!black, mark=*, mark size=1.8pt, line width=1.8pt, forget plot]
    table[x=step, y=glm4_9b_noisy_mean] {\curvedata};
\addlegendimage{
    legend image code/.code={
        \draw[blue!75!black, line width=1.8pt] (0cm,0cm) -- (0.5cm,0cm);
        \fill[blue!75!black] (0.25cm,0cm) circle (1.6pt);
    }
}
\addlegendentry{Group Rollout $p{=}0.10$}

\nextgroupplot[
    title={Qwen3-8B},
    legend style={at={(0.97,0.97)},anchor=north east,font=\footnotesize,draw=black,fill=white},
]
\addplot[name path=qwenBlUpper, draw=none, forget plot] table[x=step, y expr=\thisrow{qwen3_8b_baseline_mean}+\thisrow{qwen3_8b_baseline_std}] {\curvedata};
\addplot[name path=qwenBlLower, draw=none, forget plot] table[x=step, y expr=\thisrow{qwen3_8b_baseline_mean}-\thisrow{qwen3_8b_baseline_std}] {\curvedata};
\addplot[black!30, fill opacity=0.50, forget plot] fill between[of=qwenBlUpper and qwenBlLower];
\addplot[color=black, mark=square*, mark size=1.6pt, line width=1.8pt, forget plot]
    table[x=step, y=qwen3_8b_baseline_mean] {\curvedata};
\addlegendimage{
    legend image code/.code={
        \draw[black, line width=1.8pt] (0cm,0cm) -- (0.5cm,0cm);
        \fill[black] (0.25cm,0cm) +(-1.4pt,-1.4pt) rectangle +(1.4pt,1.4pt);
    }
}
\addlegendentry{No Noise}
\addplot[name path=qwenNUpper, draw=none, forget plot] table[x=step, y expr=\thisrow{qwen3_8b_noisy_mean}+\thisrow{qwen3_8b_noisy_std}] {\curvedata};
\addplot[name path=qwenNLower, draw=none, forget plot] table[x=step, y expr=\thisrow{qwen3_8b_noisy_mean}-\thisrow{qwen3_8b_noisy_std}] {\curvedata};
\addplot[orange!45, fill opacity=0.50, forget plot] fill between[of=qwenNUpper and qwenNLower];
\addplot[color=orange!90!black, mark=*, mark size=1.8pt, line width=1.8pt, forget plot]
    table[x=step, y=qwen3_8b_noisy_mean] {\curvedata};
\addlegendimage{
    legend image code/.code={
        \draw[orange!90!black, line width=1.8pt] (0cm,0cm) -- (0.5cm,0cm);
        \fill[orange!90!black] (0.25cm,0cm) circle (1.6pt);
    }
}
\addlegendentry{Group Rollout $p{=}0.10$}

\end{groupplot}
\end{tikzpicture}
    \caption{Median response length over training steps for group rollout noise at $p{=}0.1$ compared to the no-noise baseline. Shaded regions indicate $\pm 1$ standard deviation across seeds.}
    \label{fig:response-length}
\end{figure}

However, \citet{liuUnderstandingR1ZeroLikeTraining2025} noted that output length does not imply better downstream performance. In qualitative inspection of early rollouts, we observe long, sometimes repetitive reasoning chains and responses that exhaust the token budget before producing an answer. Paragraphs often start with expressions such as ``Alternatively,'' ``But wait,'' and ``Wait.'' These observations do not establish why the models produce these patterns.

Shorter responses may help the models produce complete answers within the token budget. Missing-code penalties remain uncorrupted, so mixed groups retain an incentive to produce an extracted code block even when correctness rewards are highly noisy (\Cref{app:group-noise-analysis}). We do not quantify truncation rates, format compliance or correctness conditional on a complete answer. Consequently, these results do not separate improvements in presenting an answer from improvements in solving the task.

\section{Llama results}\label{app: llama-results}

We include a subset of experiments with Llama~3.1~8B as a check beyond the Qwen/GLM model families. \Cref{fig:llama-results} shows its group rollout noise sweep and training curves. The clean baseline reaches a best validation reward of $0.658$ under our protocol. The noisy runs' peaks are nonmonotonic across the tested rates, ranging from $0.643$ at $p{=}0.20$ to $0.678$ at $p{=}0.05$. Final scores are substantially below the peaks in both clean and noisy conditions. Each noisy setting has one seed, so these observations do not establish a beneficial effect of noise, equivalent performance or general robustness across model families.

\begin{figure}[t]
    \centering
    \begin{tikzpicture}
\colorlet{noiseA}{blue!60}          %
\colorlet{noiseB}{green!55!black}   %
\colorlet{noiseC}{orange!85}        %
\colorlet{noiseD}{red!70}           %
\colorlet{llamagreen}{green!55!black} %

\def\curveLW{2.2pt}
\def\thinLW{1.6pt}
\pgfplotsset{table/search path={.,shared/diagram/llama_noise}}

\pgfplotstableread[col sep=comma]{arxiv-shared--diagram--llama_noise--data--llama_reward_vs_noise.csv}\rewardVsNoise
\pgfplotstableread[col sep=comma]{arxiv-shared--diagram--llama_noise--data--llama_curve_baseline.csv}\curveBaseline
\pgfplotstableread[col sep=comma]{arxiv-shared--diagram--llama_noise--data--llama_curve_p005.csv}\curvePfive
\pgfplotstableread[col sep=comma]{arxiv-shared--diagram--llama_noise--data--llama_curve_p010.csv}\curvePten
\pgfplotstableread[col sep=comma]{arxiv-shared--diagram--llama_noise--data--llama_curve_p015.csv}\curvePfifteen
\pgfplotstableread[col sep=comma]{arxiv-shared--diagram--llama_noise--data--llama_curve_p020.csv}\curvePtwenty
\pgfplotstableread[col sep=comma]{arxiv-shared--diagram--llama_noise--data--llama_resplen_baseline.csv}\resplenBaseline
\pgfplotstableread[col sep=comma]{arxiv-shared--diagram--llama_noise--data--llama_resplen_noisy.csv}\resplenNoisy

\def\baselinebestmean{0.6584}
\def\baselinefinalmean{0.5048}
\def\basemodelreward{0.5967}
\def\basemodelresplen{341.2}

\begin{groupplot}[
    group style={
        group size=3 by 1,
        horizontal sep=1.6cm,
    },
    scale only axis,
    width=0.225\columnwidth,
    height=0.225\columnwidth,
    ymajorgrids=true,
    grid style={dashed,gray!35},
    tick label style={font=\footnotesize},
]

\nextgroupplot[
    xlabel={Noise level ($p$)},
    ylabel={Reward},
    xmin=0.03, xmax=0.22,
    ymin=0.45, ymax=0.72,
    ytick={0.45,0.50,0.55,0.60,0.65,0.70},
    yticklabels={0.45,0.50,0.55,0.60,0.65,0.70},
    xtick={0.05,0.10,0.15,0.20},
    xticklabels={0.05,0.10,0.15,0.20},
]
\addplot[color=llamagreen, mark=*, mark size=2pt, line width=\curveLW, forget plot]
    table[x=noise_p, y=best_mean] {\rewardVsNoise};
\addplot[color=llamagreen, mark=square*, mark size=1.8pt, line width=\thinLW, dashed, opacity=0.7, forget plot]
    table[x=noise_p, y=final_mean] {\rewardVsNoise};
\addplot[black, line width=\thinLW, forget plot]
    coordinates {(0.03,\baselinebestmean) (0.22,\baselinebestmean)};
\addplot[black, dashdotted, line width=\thinLW, opacity=0.5, forget plot]
    coordinates {(0.03,\baselinefinalmean) (0.22,\baselinefinalmean)};
\addplot[black, dotted, line width=\thinLW, opacity=0.7, forget plot]
    coordinates {(0.03,\basemodelreward) (0.22,\basemodelreward)};

\nextgroupplot[
    xlabel={Steps},
    ylabel={Reward},
    xmin=0, xmax=260,
    ymin=0.45, ymax=0.72,
    ytick={0.45,0.50,0.55,0.60,0.65,0.70},
    yticklabels={0.45,0.50,0.55,0.60,0.65,0.70},
    xtick={0,50,100,150,200,250},
]
\addplot[color=black, mark=none, line width=\curveLW, forget plot]
    table[x=step, y=mean] {\curveBaseline};
\addplot[name path=blup, draw=none, forget plot]
    table[x=step, y expr=\thisrow{mean}+\thisrow{std}] {\curveBaseline};
\addplot[name path=bllo, draw=none, forget plot]
    table[x=step, y expr=\thisrow{mean}-\thisrow{std}] {\curveBaseline};
\addplot[fill=black, opacity=0.08, forget plot] fill between[of=blup and bllo];
\addplot[color=noiseA, mark=none, line width=\thinLW, forget plot]
    table[x=step, y=mean] {\curvePfive};
\addplot[color=noiseB, mark=none, line width=\thinLW, forget plot]
    table[x=step, y=mean] {\curvePten};
\addplot[color=noiseC, mark=none, line width=\thinLW, forget plot]
    table[x=step, y=mean] {\curvePfifteen};
\addplot[color=noiseD, mark=none, line width=\thinLW, forget plot]
    table[x=step, y=mean] {\curvePtwenty};
\addplot[black, dotted, line width=\thinLW, opacity=0.5, forget plot]
    coordinates {(0,\basemodelreward) (260,\basemodelreward)};

\nextgroupplot[
    xlabel={Steps},
    ylabel={Median response length},
    xmin=0, xmax=260,
    xtick={0,50,100,150,200,250},
    scaled y ticks=false,
    yticklabel={\pgfmathparse{\tick/1000}\pgfmathprintnumber[fixed,precision=0]{\pgfmathresult}k},
]
\addplot[color=black, mark=none, line width=\curveLW, forget plot]
    table[x=step, y=mean] {\resplenBaseline};
\addplot[name path=rlblup, draw=none, forget plot]
    table[x=step, y expr=\thisrow{mean}+\thisrow{std}] {\resplenBaseline};
\addplot[name path=rlbllo, draw=none, forget plot]
    table[x=step, y expr=\thisrow{mean}-\thisrow{std}] {\resplenBaseline};
\addplot[fill=black, opacity=0.1, forget plot] fill between[of=rlblup and rlbllo];
\addplot[color=llamagreen, mark=none, line width=\curveLW, forget plot]
    table[x=step, y=mean] {\resplenNoisy};
\addplot[name path=rlnup, draw=none, forget plot]
    table[x=step, y expr=\thisrow{mean}+\thisrow{std}] {\resplenNoisy};
\addplot[name path=rlnlo, draw=none, forget plot]
    table[x=step, y expr=\thisrow{mean}-\thisrow{std}] {\resplenNoisy};
\addplot[fill=llamagreen, opacity=0.15, forget plot] fill between[of=rlnup and rlnlo];
\addplot[black, dotted, line width=\thinLW, opacity=0.5, forget plot]
    coordinates {(0,\basemodelresplen) (260,\basemodelresplen)};

\end{groupplot}

\node[anchor=north, inner sep=0pt, yshift=-1.2cm, font=\footnotesize] at (group c2r1.south) {
    \setlength{\tabcolsep}{3pt}%
    \begin{tabular}{@{}
        r@{\,}l @{\quad} r@{\,}l @{\quad} r@{\,}l @{\quad} r@{\,}l @{\quad} r@{\,}l
    @{}}
        \tikz{\draw[llamagreen,line width=\curveLW](0,0)--(0.4,0); \fill[llamagreen](0.2,0) circle(1.3pt);} & Best &
        \tikz{\draw[llamagreen,line width=\thinLW,dashed,opacity=0.7](0,0)--(0.4,0);} & Final &
        \tikz{\draw[black,line width=\curveLW](0,0)--(0.4,0);} & Baseline &
        \tikz{\draw[llamagreen,line width=\curveLW](0,0)--(0.4,0);} & Noisy ($p{=}0.10$) &
        \tikz{\draw[black,dotted,line width=\thinLW](0,0)--(0.4,0);} & Base model \\[2pt]
        \tikz{\draw[noiseA,line width=\thinLW](0,0)--(0.4,0);} & $p{=}0.05$ &
        \tikz{\draw[noiseB,line width=\thinLW](0,0)--(0.4,0);} & $p{=}0.10$ &
        \tikz{\draw[noiseC,line width=\thinLW](0,0)--(0.4,0);} & $p{=}0.15$ &
        \tikz{\draw[noiseD,line width=\thinLW](0,0)--(0.4,0);} & $p{=}0.20$ \\
    \end{tabular}
};
\end{tikzpicture}
    \caption{Llama~3.1~8B with group rollout noise on MBPP. \textbf{Left:} best and final validation reward vs.\ noise level. \textbf{Center:} training curves by noise level. \textbf{Right:} median response length comparison.}
    \label{fig:llama-results}
\end{figure}

\section{Cross-study evaluation comparisons}\label{sec:cai protocol comparison}
\citet{caiReinforcementLearningVerifiable2025} report mathematical reasoning results under clean and noisy verifier rewards, including moderate asymmetric noise, and compare reward-correction methods. Their results are relevant to our study. Different outcomes across studies can reflect different error structures, training procedures and evaluation protocols; they do not by themselves establish a contradiction.

\Cref{tab:cai cross-study comparison} collects previously tabulated comparisons with the Qwen2.5-Math and DeepSeek-R1 technical reports \citep{yangQwen25MathTechnicalReport2024,deepseek-aiDeepSeekR1IncentivizingReasoning2025}. The ranges run from the reported base-model scores to the scores after training with clean rewards; they are not uncertainty intervals. Among the 12 populated comparisons, five reference scores exceed the upper endpoint, including one difference of only $0.2$ percentage points at the displayed precision.

We have not matched prompts, decoding settings, response-length budgets, answer extraction or benchmark variants across these sources. The table therefore cannot diagnose invalid evaluation or show that training worsens a model relative to its own baseline. Our Qwen3 baseline also differs from its technical report under our protocol, as noted above. A valid comparison of training effects uses a common protocol within each study; the cross-study discrepancies are a comparability limitation rather than grounds for excluding prior findings.

\begin{table}[t]
    \centering\small
    \caption{Previously tabulated pass@1 comparisons for Qwen2.5-Math-1.5B, Qwen2.5-Math-7B and DeepSeek-R1-Distill-Qwen-1.5B. References are the corresponding technical reports \citep{yangQwen25MathTechnicalReport2024,deepseek-aiDeepSeekR1IncentivizingReasoning2025}; ranges span the base and clean-reward post-training scores in Figure~2 of \citet{caiReinforcementLearningVerifiable2025}. Evaluation protocols and benchmark variants have not been matched, so these numbers do not establish inconsistent evaluation or negative training effects.}
    \label{tab:cai cross-study comparison}
    \vspace{4pt}
    \setlength{\tabcolsep}{2pt}
    \begin{tabular}{c|c c|c c|c c}
    \toprule
    & \multicolumn{2}{c|}{Qwen2.5-Math-1.5B} & \multicolumn{2}{c|}{Qwen2.5-Math-7B} & \multicolumn{2}{c}{DeepSeek-R1-Distill-Qwen-1.5B} \\
                        & Reference & Range & Reference & Range & Reference & Range \\\midrule
        Math            & $69.4$ & $[48, 70]$ & $75.1$ & $[52, 80]$ & $83.9$ & $[60, 78]$ \\
        MinervaMath     & $29.4$ & $[6, 18]$  & $34.6$ & $[10, 25]$ & & \\
        Olympiad Bench  & $31.3$ & $[24, 32]$ & $38.2$ & $[26, 38]$ & & \\
        AIME 2024       & $3.3$  & $[7, 16]$  & $13.3$ & $[12, 30]$ & $28.9$ & $[10, 20]$\\
        AMC 2023        & $45.0$ & $[34, 50]$ & $62.5$ & $[45, 63]$ & & \\
        \bottomrule
    \end{tabular}
\end{table}

\section{Ackley function optimization}
\label{app: ackley function optimization}

\subsection{Setup}

This illustration uses additive perturbations of scalar rewards, not the whole-group outcome complementation analyzed in \Cref{app:group-noise-analysis}. The available archive does not recover the numerical policy standard deviation, group size, learning rate, initialization radius, or the distribution and across-response dependence of the additive noise. The original executable implementation is also unavailable in the archive. The equations below describe the recorded update construction but are insufficient to reproduce the plotted trajectories exactly.

The ``policy'' is a diagonal Gaussian with fixed standard deviation:
\[
  \pi(\cdot \mid \mu) = \mathcal{N}(\mu,\, \sigma^2 I),
  \qquad \mu \in \mathbb{R}^2.
\]
$\mu$ is the only learnable parameter, optimized by Adam \citep{kingma2015adam}.

\subsection{Forward pass (one iteration)}

\subsubsection{Sampling}

Samples are generated with $\mu$ \emph{detached} from the computation graph:
\[
  s_i = \mu_{\mathrm{detach}} + \sigma\,\varepsilon_i,
  \qquad \varepsilon_i \sim \mathcal{N}(0, I),
  \quad i = 1,\dots,G.
\]
Because $\mu$ is detached, the samples $s_i$ are constants with no gradient connection to $\mu$.

\subsubsection{Rewards and advantages}

The Ackley cost is evaluated on these constant samples and converted to group-relative advantages:
\[
  r_i = -\operatorname{ackley}(s_i) + \eta_i,
  \qquad
  A_i = \frac{r_i - \bar{r}}{\operatorname{std}(r) + 10^{-8}}.
\]
Here $\eta_i$ denotes the additive reward perturbation; its dependence across responses is not specified in the available records. A perturbation shared identically by every response would cancel during group centering. There are no gradients through the rewards or advantages.

\subsubsection{Log-probabilities (where gradients enter)}

A distribution $\mathcal{N}(\mu, \sigma^2 I)$ is constructed with the \emph{live} $\mu$, and log-probabilities are evaluated on the constant samples:
\[
  \log \pi(s_i \mid \mu)
  = \sum_{d=1}^{2} \left[
      -\frac{(s_{i,d} - \mu_d)^2}{2\sigma^2}
      - \log\sigma
      - \tfrac{1}{2}\log 2\pi
    \right].
\]
The above expression is the only point where $\mu$ enters the computation graph.

\subsubsection{Policy loss}

The loss is the standard REINFORCE objective with group-relative advantages:
\[
  L = -\frac{1}{G}\sum_{i=1}^{G} A_i \cdot \log\pi(s_i \mid \mu).
\]

Differentiating the log-probability with respect to~$\mu_d$:
\[
  \frac{\partial \log\pi(s_i \mid \mu)}{\partial \mu_d} = \frac{s_{i,d} - \mu_d}{\sigma^2}.
\]

Since $s_{i,d} - \mu_d = \sigma\,\varepsilon_{i,d}$ (the detached and live
$\mu$ values share the same value at evaluation time):
\[
  \frac{\partial \log\pi(s_i \mid \mu)}{\partial \mu_d} = \frac{\varepsilon_{i,d}}{\sigma}.
\]

\subsection{Gradient}

\[
    \nabla_\mu L = -\frac{1}{G\,\sigma}\sum_{i=1}^{G} A_i\,\varepsilon_i
\]

Each step pushes $\mu$ toward samples that received above-average rewards and
away from below-average ones, weighted by the normalized noise directions.
Adam applies adaptive learning rates on top of this raw gradient.
 
\end{document}